\documentclass{article}

\usepackage[preprint]{neurips_2024}

\usepackage{amsmath}
\usepackage{amssymb}
\usepackage{mathtools}
\usepackage{amsthm}
\usepackage{natbib}
\setcitestyle{numbers,square}
\usepackage{multirow}
\usepackage[ruled]{algorithm2e}
\usepackage{bbding}

\usepackage[utf8]{inputenc} % allow utf-8 input
\usepackage[T1]{fontenc}    % use 8-bit T1 fonts
\usepackage{url}            % simple URL typesetting
\usepackage{booktabs}       % professional-quality tables
\usepackage{amsfonts}       % blackboard math symbols
\usepackage{nicefrac}       % compact symbols for 1/2, etc.
\usepackage{microtype}      
\usepackage[dvipsnames,cmyk,table]{xcolor}        
\usepackage{listings}

\usepackage{arydshln}
\usepackage{xspace}
\usepackage{float}
\usepackage{wrapfig}
\usepackage{lipsum}

\usepackage{color, soul}
\usepackage{colortbl}
\usepackage{setspace}

\usepackage{hyperref}       

\hypersetup{
  colorlinks=true,
  citecolor=[rgb]{0,0.55,0.27},
  urlcolor=RoyalBlue,
  linkcolor=[rgb]{0,0.55,0.27}
}

\title{Enhancing LLM Reasoning via Critique Models with Test-Time and Training-Time Supervision}

\author{
Zhiheng Xi$^1$\thanks{Equal contribution.}\ \ \thanks{Correspondence to: zhxi22@m.fudan.edu.cn, tgui@fudan.edu.cn } , Dingwen Yang$^1$$^*$, Jixuan Huang$^1$, Jiafu Tang$^1$, Guanyu Li$^1$, Yiwen Ding$^1$,\\
\textbf{Wei He$^1$, Boyang Hong$^1$, Shihan Dou$^1$, Wenyu Zhan$^1$, Xiao Wang$^1$, Rui Zheng$^1$, Tao Ji$^1$,}\\
\textbf{Xiaowei Shi$^2$, Yitao Zhai$^2$, Rongxiang Weng$^2$, Jingang Wang$^2$, Xunliang Cai$^2$, }\\
\textbf{Tao Gui$^1$$^\dag$, Zuxuan Wu$^1$, Qi Zhang$^1$, Xipeng Qiu$^1$, Xuanjing Huang$^1$, Yu-Gang Jiang$^1$}\\ \\
\large\text{$^1$Fudan University $^2$Meituan \  \ \ \ \ { }{ }}
}

\definecolor{lightgray}{RGB}{211, 211, 211}
\definecolor{lightfont}{gray}{0.3}

\begin{document}

\maketitle

\begin{abstract}

Training large language models (LLMs) to spend more time thinking and reflection before responding is crucial for effectively solving complex reasoning tasks in fields such as science, coding, and mathematics. However, the effectiveness of mechanisms like self-reflection and self-correction depends on the model's capacity to accurately assess its own performance, which can be limited by factors such as initial accuracy, question difficulty, and the lack of external feedback. In this paper, we delve into a two-player paradigm that separates the roles of reasoning and critique models, where the critique model provides step-level feedback to supervise the reasoning (actor) model during both test-time and training-time. We first propose AutoMathCritique, an automated and scalable framework for collecting critique data, resulting in a dataset of $76,321$ responses paired with step-level feedback. Fine-tuning language models with this dataset enables them to generate natural language feedback for mathematical reasoning. We demonstrate that the critique models consistently improve the actor's performance on difficult queries at test-time, especially when scaling up inference-time computation. Motivated by these findings, we introduce the critique-based supervision to the actor's self-training process, and propose a critique-in-the-loop self-improvement method. Experiments show that the method improves the actor's exploration efficiency and solution diversity, especially on challenging queries, leading to a stronger reasoning model. Lastly, we take the preliminary step to explore training self-talk reasoning models via critique supervision and showcase their potential. Our code and datasets are at \href{https://mathcritique.github.io/}{https://mathcritique.github.io/}.

\end{abstract}

\section{Introduction}

With the rapid advancement of large language models (LLMs) \citep{DBLP:conf/nips/Ouyang0JAWMZASR22,DBLP:journals/corr/abs-2303-08774,DBLP:journals/corr/abs-2307-09288,DBLP:journals/corr/abs-2310-06825,DBLP:journals/corr/abs-2407-21783}, significant progress has been made in enhancing their reasoning capabilities \citep{DBLP:conf/nips/Wei0SBIXCLZ22,DBLP:conf/emnlp/XiJZZGLGZH23,DBLP:journals/corr/abs-2308-08998,DBLP:conf/nips/YaoYZS00N23,DBLP:journals/corr/abs-2308-09583,DBLP:journals/corr/abs-2402-05808}. By prompting or training language models to reason step-by-step like humans (i.e., chain-of-thought, CoT), these models have demonstrated impressive reasoning abilities \citep{DBLP:conf/nips/Wei0SBIXCLZ22,DBLP:conf/nips/YaoYZS00N23,DBLP:conf/iclr/YaoZYDSN023}. Recently, OpenAI's o1 model has introduced a new paradigm shift, exploring to increase inference-time computation in language models and explicitly generate longer chains of thought \citep{openai2024learning}. This enables them to tackle more complex reasoning tasks that even humans find challenging, such as problems in the domains of science, coding, and mathematics \citep{DBLP:conf/nips/ShinnCGNY23,DBLP:journals/corr/abs-2408-03314,DBLP:journals/corr/abs-2407-18219,DBLP:journals/corr/abs-2409-12917}.

At the same time, many studies have explored test-time scaling by employing mechanisms like self-reflection, self-correction, and self-critique to generate longer thinking chains \citep{DBLP:conf/iclr/WelleckLWBSK023,DBLP:conf/nips/ShinnCGNY23,DBLP:conf/acl/XuZZP0024,DBLP:conf/iclr/YaoZYDSN023,selfee2023,DBLP:conf/nips/KimBM23}, similar to OpenAI's o1. However, the effectiveness of these mechanisms depends on the models' ability to accurately evaluate their own performance. This ability can be limited by factors such as initial accuracy, problem complexity, and the lack of external feedback \citep{DBLP:journals/corr/abs-2409-12917,DBLP:journals/corr/abs-2206-05802,DBLP:conf/nips/MadaanTGHGW0DPY23,DBLP:conf/iclr/WelleckLWBSK023}. As a result, their performance remains constrained, even with increased inference-time computation \citep{DBLP:conf/iclr/0009CMZYSZ24}.

In light of this, to reliably increase reasoning models' performance with increased inference-time computation, we delve into a two-player paradigm, where the actor model engages in reasoning while the critique model provides supervisory feedback on the thought chains \citep{DBLP:conf/iclr/WelleckLWBSK023,DBLP:conf/acl/AkyurekAKCWT23,DBLP:conf/iclr/YaoHNLFXNC0AXMW24,DBLP:conf/icml/HavrillaRNDZHR24}. This approach represents a scalable oversight technique aiming at providing reliable and effective supervision for the continued development of LLMs \citep{DBLP:journals/corr/abs-2206-05802,bai2022constitutional,DBLP:journals/corr/abs-2211-03540}. The goal is to help the actor model identify errors and refine its outputs, ultimately leading to higher-quality results. In this paper, We aim to explore the research question of how to develop effective and reliable critique models, and how to enhance the actor’s reasoning performance through collaboration with the critique model at test-time. Additionally, we explore incorporating supervision from critique models into the actor's training process to build more capable reasoning models.

\begin{figure*}[t]
    \includegraphics[width=0.9\linewidth]{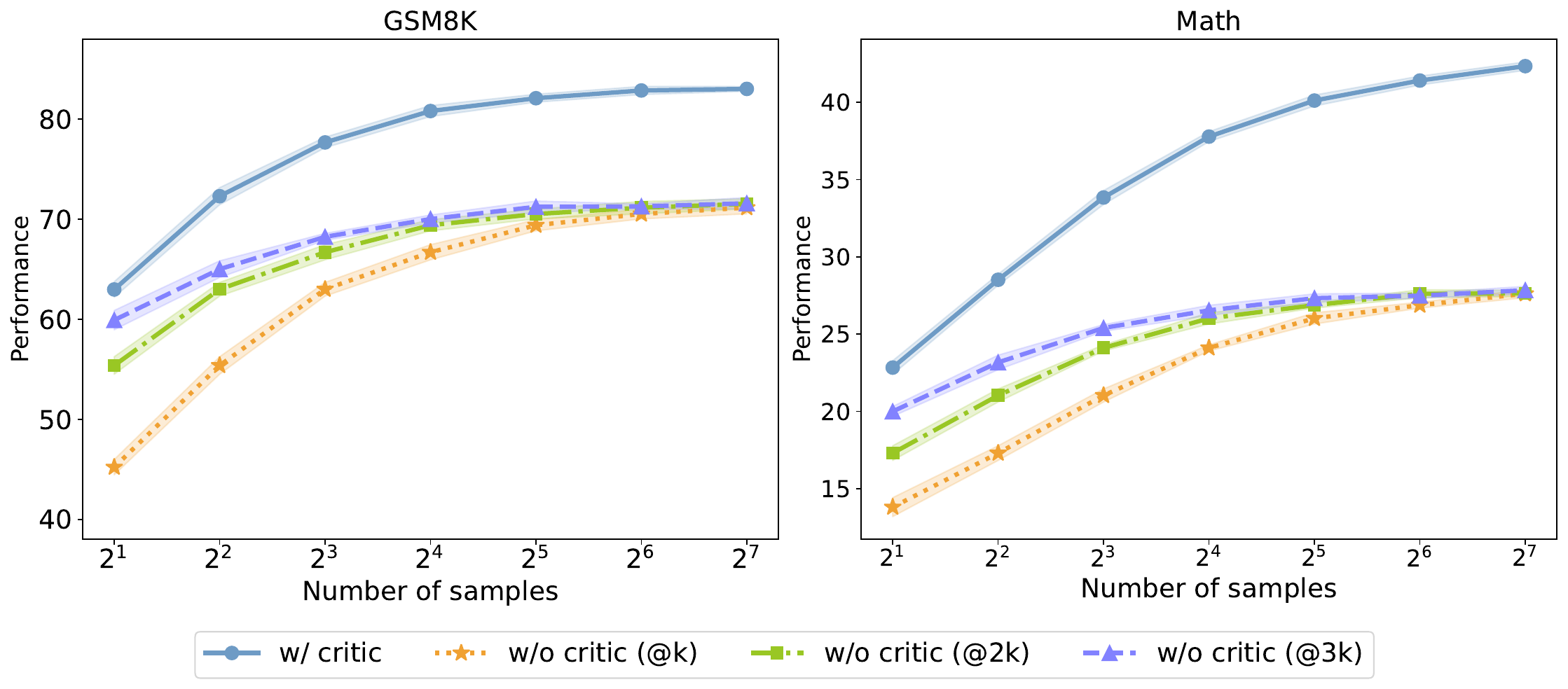}
    \centering
    \vspace{-1pt}
 	\caption{
    Majority voting (Maj@K) performance when scaling test-time computation. The x-axis represents the number of samples, and the y-axis represents performance. ``$2K$'' and ``$3K$'' denote using $2\times$ and $3\times$ the sampling amount shown on the x-axis, respectively. Without critique models, Maj@K performance quickly plateaus as computation increases. In contrast, using critique models consistently improves the performance ceiling by a significant margin.
}\label{fig:MV_test_time_scaling}
    % \vspace{-8pt}
\end{figure*}

\begin{figure*}[t]
    \includegraphics[width=0.9\linewidth]{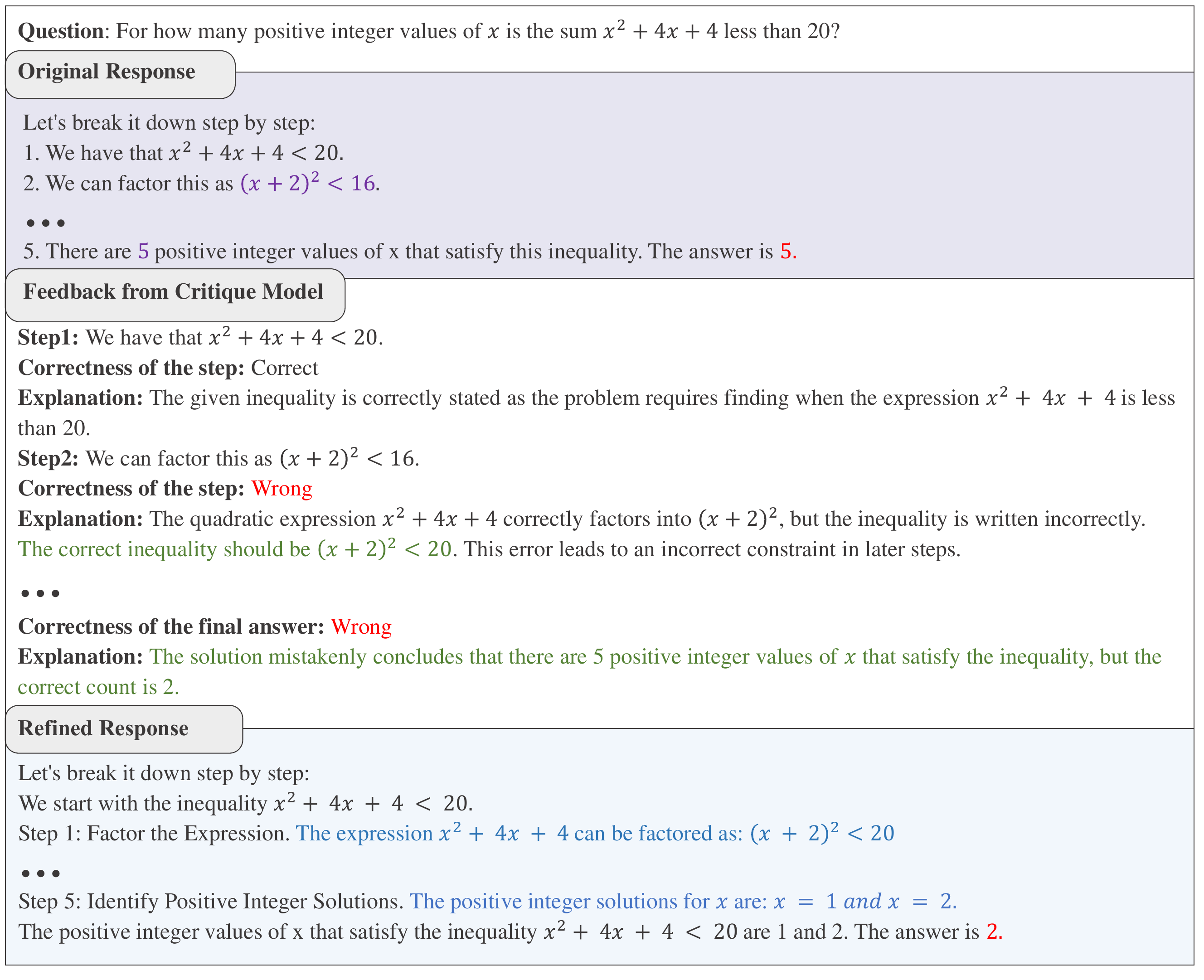}
    \centering
    \vspace{-0.3cm}
 	\caption{ An example of the response, critique, and refinement process in the two-player setting.
  }\label{fig:case_study_1}
  \vspace{-0.18cm}
\end{figure*}

We first propose an automated and scalable framework called \textbf{AutoMathCritique} to collect diverse and high-quality step-level critique data without additional human supervision (Section \ref{sec:AutoMathCritique: An Automated and Scalable Framework to Collect Step-level Critique Data}). The framework consists of three main stages: flawed reasoning path construction, critique generation, and data filtering. In the first step, we leverage several approaches for controlled error synthesis, each of which targets different aspects of reasoning errors, such as their location or specific content. This controlled process ensures the diversity and comprehensiveness of the reasoning paths and provides informative and precise hints to guide the subsequent critique generation. In the second step, annotator models are provided with the original reasoning path, and possible hints about the mistakes to label step-level correctness and offer constructive feedback. In the second step, the reasoning model revises the response according to the critiques, and Monte Carlo sampling \citep{DBLP:conf/acl/WangLSXDLCWS24,DBLP:journals/corr/abs-2410-08146} is used to eliminate low-quality or non-informative critique data, while preventing high-quality data from being accidentally discarded. A case of the resulting data is illustrated in Figure \ref{fig:case_study_1}.

Next, using AutoMathCritique, we create a critique dataset containing $76321$ samples named \textbf{MathCritique-76k}, which is subsequently used to fine-tune a language model to obtain the critique model. We demonstrate that the critique models can assist the actor model in improving exploration efficiency and reasoning quality during test time, leading to a significant enhancement in its reasoning performance (Section \ref{sec:Critique Models Improves LLM Reasoning through Test-time Supervision}). Through in-depth analysis, we find that the critique models are particularly effective in helping the actor achieve better results on difficult queries. Additionally, by scaling inference-time computation \citep{DBLP:journals/corr/abs-2408-03314,DBLP:journals/corr/abs-2407-21787,DBLP:journals/corr/abs-2408-16737}, the performance gains brought by the critique models continue to grow. 

Motivated by the insights of test-time, we introduce the critique model into the actor model's exploration and learning process, introducing a \textbf{critique-in-the-loop self-improvement} method (Section \ref{sec: Critique-in-the-loop Self-Improvement for Better Reasoning Models}). With the supervision of critique models and by scaling exploration computation for difficult queries, our method improves the actor's exploration efficiency and solution diversity, alleviating the issue of tail narrowing \citep{ding2024mitigating} in reasoning models during iterative exploration and learning. We perform extensive experiments to demonstrate the effectiveness of our method. Additionally, we conduct further analysis of the critique models (Section \ref{sec:Discussion and Analysis}), e.g., the scaling properties, and whether we should scale test-time computation in sequential or parallel.

Finally, we take a step further and conduct preliminary explorations on how to leverage critique data to construct step-level self-talk data (Section \ref{sec:A Step Further: Training Step-level Self-Talk Reasoning Models via Critique Data}). We propose the \textbf{self-talk-via-critique} method, and train a single language model to reflect and self-correct at each step, demonstrating the potential of this approach.

In summary, our main contributions are: 
\begin{itemize}
    \item We introduce AutoMathCritique, an automated and scalable framework for collecting step-level critique data without additional human supervision, which we use to build the large-scale critique dataset MathCritique-76k.
    \item We fine-tune the critique model with MathCritique-76k to offer constructive feedback on reasoning paths. We demonstrate and analyze the performance gains of the trained critique models in enhancing the actor's reasoning during test time, particularly when scaling test-time computation.
    \item Motivated by the insights from test-time analysis, we introduce the critique model to the actor's self-training process, and propose the critique-in-the-loop self-improvement method to enhance exploration efficiency and solution diversity, ultimately training better reasoning models.
    \item We conduct extensive experiments to validate the effectiveness of our method and perform in-depth analysis of critique models, e.g., their scaling properties, and whether we should scale test-time computation in sequential or parallel.
    \item We propose the self-talk-via-critique method, and take the preliminary step to train models that can perform step-level reasoning, reflection and correction, and demonstrate their potential. We hope our work offers valuable insights for future research on LLM reasoning and scalable supervision.
\end{itemize}

\section{Preliminaries}

In the two-player setting studied in this paper, there are two roles: the actor model and the critique model. Also, there are three primary tasks \citep{DBLP:journals/corr/abs-2206-05802}: reasoning, critique, and refinement. 

In the reasoning task, the actor model $\pi_{\theta}$ parameterized by $\theta$ is given a reasoning problem $x$ and is expected to generate a response $y = \pi_{\theta}(x)$. This response includes both the answer to the problem and the reasoning trajectory. The accuracy of this response can be evaluated using a reward function $r(x,y)$.

Next, the critique model $\pi_{\phi}$ parameterized by $\pi_{\phi}$ performs the critique task, where, given the problem and response, it generates critical feedback $c = \pi_{\phi}(x,y)$. Notably, if the oracle reward function of the response is not given, the critique task consists of two subtasks: the discriminative task and the feedback generation task. The former determines whether the response contains flaws, while the latter generates constructive natural language feedback. 

Finally, we define the refinement task, in which, given the problem, response, and critique, the actor generates a new response $y' = \pi_{\theta}(x,y,c)$—this is also known as conditional refinement. Alternatively, we can define direct refinement $y' = \pi_{\theta}(x,y)$, where the actor provides an improved answer based on an existing answer without conditioning on a critique, which is also referred to as ``self-correction'' \citep{DBLP:conf/iclr/WelleckLWBSK023}.

This process can proceed in multiple rounds. We define that in the initial round (round $0$) only the actor operates, generating a response based on the problem. In round \(i\), the critique model first generates a new critique based on the interaction history, which is represented as:
\[ 
c_i = \pi_{\phi}(x, y_0, c_1, y_1, ..., c_{i-1}, y_{i-1}) .
\]
Then, the actor generates a new refinement based on the previous interaction history, represented as:
\[ 
y_i = \pi_{\theta}(x, y_0, c_0, y_1, c_1, ..., c_{i-1}).
\]

\section{AutoMathCritique: An Automated and Scalable Framework to Collect Step-level Critique Data}
\label{sec:AutoMathCritique: An Automated and Scalable Framework to Collect Step-level Critique Data}

\begin{figure}[t]
    \begin{center}
    \includegraphics[width=0.96\textwidth]
    {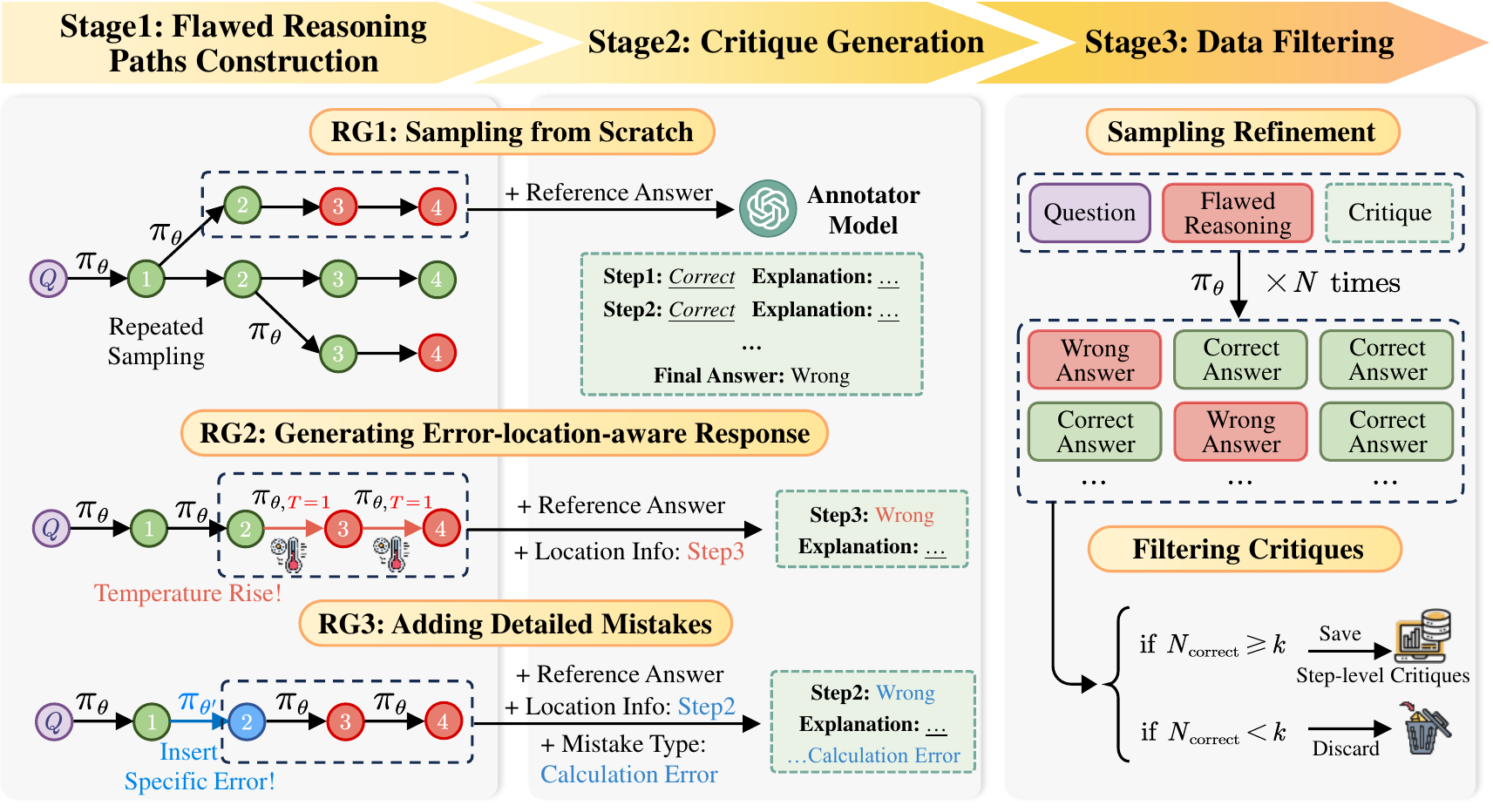} 
    \caption{The overview of AutoMathCritique framework. It has three main steps: flawed reasoning path construction, critique generation, and data filtering.}
    \label{fig:AutoMathCritique}
    \end{center}
\end{figure}

To train critique models capable of delivering step-level supervision and constructive feedback for reasoning, we introduce AutoMathCritique—an automated and scalable framework for collecting critique data (see Figure \ref{fig:AutoMathCritique} for an overview of AutoMathCritique).
This framework consists of three main stages: flawed reasoning path construction, critique generation, and data filtering. Using AutoMathCritique, we create a dataset containing $76,321$ samples named MathCritique-76k. The statistics are listed in Table \ref{tab:dataset_statistics}.

We focus on the field of mathematical reasoning, so we utilize two of the most widely used datasets: GSM8K \cite{DBLP:journals/corr/abs-2110-14168} and MATH \cite{DBLP:conf/nips/HendrycksBKABTS21}. The queries used for our subsequent data construction primarily come from their training sets, and we also leverage their original annotated responses to train the actor reasoning models. Our in-domain test set is composed of their test sets.

\subsection{Construction of Flawed Reasoning Paths}

To create high-quality critique data, we first need to construct a dataset of reasoning paths that includes some flaws. To better control the quality and diversity of the generated flawed reasoning paths, and to facilitate the subsequent construction of critique data, we leverage several distinct response generation (RG) approaches. These strategies encompass different aspects of the errors, such as their location or specific details. We mainly use Llama3-8B \cite{DBLP:journals/corr/abs-2407-21783} as our actor model for sampling.

\paragraph{RG1: sampling from scratch.}
In this approach, the actor is provided with a query and tasked with generating a response. Given that the actor we used has already achieved high accuracy on the GSM8K and MATH training sets, we use repeated sampling to obtain flawed responses. However, this method has the limitation of not offering detailed information about the location or content of the mistakes, which means that the subsequent critique labeling heavily depends on the expertise of annotators. 
\paragraph{RG2: generating error-location-aware response.}

In this approach, given a query, we first sample a correct response from the actor model. Then, starting from a specific step of the response, we modify the model's hyperparameters for flawed response sampling, such as increasing the temperature of the final softmax function. This ensures that the steps preceding the selected step remain consistent with the original correct response, while the subsequent steps are more likely to contain errors. If the sampled response remains correct, we select a different step and further increase the randomness of the generation process. This method strikes a balance between generating flawed responses and maintaining the coherence of the reasoning process. The correct responses we sample are later used to construct critiques, while for the flawed responses, we collect information about the error locations (e.g., identifying from which step the errors originate), thereby facilitating the annotation of high-quality critiques.

\paragraph{RG3: adding detailed mistakes.}

In this approach, given a query, the actor model is instructed to sample a correct reasoning path first. We then instruct the model to introduce mistakes into the correct response. Inspired by previous work \citep{DBLP:journals/corr/abs-2307-13702,DBLP:conf/acl/Li0LG0F24}, we enumerate various common reasoning errors in the instructions and include few-shot examples in the prompt. Each example consists of five components: the query, the correct reference response, the step where the error is introduced, the type of error, and the generated flawed response. After the error is inserted, we direct the model to continue reasoning from the erroneous step until it reaches a final answer. If a flawed response is not generated, we repeat the sampling process up to a maximum of 16 attempts. As in RG2, the correct answers obtained during this process can also be used to construct critiques. This approach allows us to easily capture information about the location of the first mistake and its specific details, thereby significantly reducing the complexity of subsequent critique construction. 

\begin{table*}[t]
\centering
\caption{Statistics of MathCritique-76k.}
\label{tab:dataset_statistics}
\vspace{2pt}
\resizebox{0.6\textwidth}{!}{ 
\begin{tabular}{lccccc}
\toprule
\multirow{1}{*}{\textbf{Dataset}} & \multirow{1}{*}{\textbf{Query}} & \multirow{1}{*}{\textbf{Golden reasoning path}}  & \multirow{1}{*}{\textbf{Critique}}  \\ 
\midrule
GSM8K   & $7,473$ & $7,473$  & $37,873$  \\
MATH    & $7,500$ & $7,498$  & $38,448$  \\
Total   & $15,973$ & $15,971$  & $76,321$  \\
\bottomrule
\end{tabular}
}
\end{table*}

\subsection{Generation of Critiques}

\paragraph{Step-level critique generation.}

When generating critique data, we enhance quality by checking each step to identify the first error in the solution, which in turn facilitates the refinement process.
Specifically, given a query and response, we employ two methods to generate step-level critique data: (1) We instruct the critique annotator (in our work, GPT-4o \citep{DBLP:journals/corr/abs-2303-08774}) to directly identify the location of the first error and provide corresponding feedback. This method requires the annotator to assess the entire solution holistically, making it relatively more challenging. (2) We instruct the annotator model to later step by step, stopping the process once the first error is detected, at which point they provide the corresponding feedback. This strategy effectively decomposes the entire solution, reducing the difficulty of providing comments.

\paragraph{Critique generation based on varying information about errors.}

When constructing responses, we employ different strategies that provide various types of information, helping annotators identify and analyze flaws. Such information plays a crucial role in generating critiques. 

For responses that are correct, we do not provide any additional information but instead ask the annotator to critique step by step. Only when the critique annotator correctly labels every step will this critique data be collected. If the annotator makes an error in labeling, it indicates either the response is a false positive (i.e., the answer is correct but the reasoning process is flawed) or the annotator's labeling is incorrect. In either case, the data is discarded.

For flawed responses, we design critique prompts based on the generation strategy used (RG1, RG2, RG3). For responses generated by RG1, we provide a correct reference response to directly assist the annotator in labeling. For flawed responses from RG2, we offer both the reference response and highlight the likely starting point of the error, helping the annotator identify the first critical mistake. For RG3-generated flawed responses, we not only specify the exact location of the error but also provide detailed information about the mistake, enabling a more precise critique.

\subsection{Data Filtering}

Although we have constructed a large amount of critique data paired with flawed responses, the quality of this data is not guaranteed, and low-quality data could weaken the performance of the critique model. 
To address this, we apply a filtering process. Specifically, we use Monte Carlo sampling: each (query, response, critique) tuple is fed into the actor model for refinement. The refinement process is repeated $10$ times, and only when the accuracy exceeds a predefined threshold $\tau=0.3$ is the critique data retained. This process is referred to as soft filtering. In contrast, hard filtering is employed when the critique is considered valid if at least one of the k refinements produces a correct result. In practice, we adopt soft filtering because it prevents the omission of high-quality critique data due to occasional model errors. Furthermore, it minimizes the risk of including low-quality critiques that the actor model does not follow, but instead refine based on its own knowledge, resulting in a correct response. 
Note that our method does not completely eliminate low-quality data, but we strive to achieve a balance between quality and quantity. Additionally, we randomly sampled $100$ data points $5$ times and had crowdsourced annotators perform the checking. We find that the rate of low-quality data is $1.2$\%.

\begin{figure*}[t]
    \includegraphics[width=0.99999\linewidth]{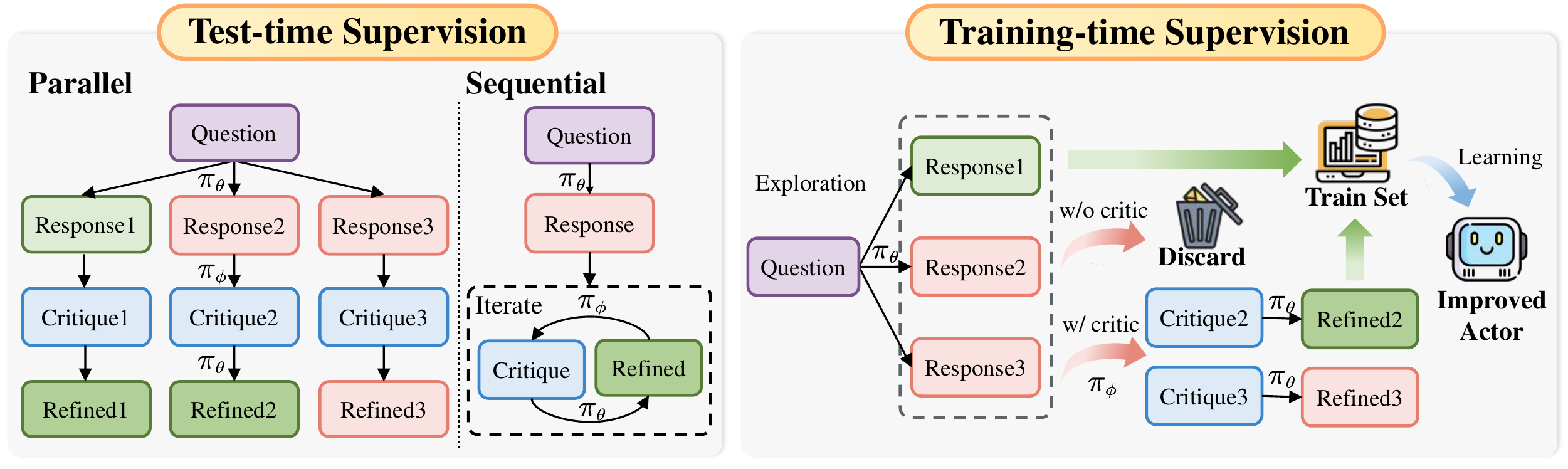}
    \centering
 	\caption{Illustration of how critique models provide supervision and constructive feedback for actor reasoning models at test-time and training-time. $\pi_{\theta}$ means the actor reasoning model, while $\pi_{\phi}$ means the critique model.}
    \label{fig: test-time-and-training-time-illustration}
\end{figure*} 

\section{Critique Models Improves LLM Reasoning through Test-time Supervision}
\label{sec:Critique Models Improves LLM Reasoning through Test-time Supervision}
In this section, we begin by training critique models to provide step-level supervisory signals and useful feedback on reasoning paths, along with the actor reasoning models that own reasoning and refinement ability (Section \ref{sec: Fine-tuning Critique Models}). We then explore the role of critique models in supporting the actor reasoning model at test-time (Section \ref{sec: Motivating Insight: Critique-in-the-loop Improves Test-time Performance}), showing that they significantly enhance the actor's performance in tackling difficult problems. Furthermore, as we scale up inference-time computations, we observe that the critique model continues to raise the performance ceiling of the reasoning models.

\subsection{Fine-tuning Critique Models and Actor Reasoning Models} \label{sec: Fine-tuning Critique Models}

\paragraph{Training critique models with MathCritique-76k.}
We train the critique models through supervised fine-tuning with the collected MathCritique-76k. Specifically, we use the standard language modeling loss. Given a dataset $\mathcal{D}_{\text{critique}}=\{x,y,c\}_{j=1}^{N}$, the loss for the critique model $\pi_{\phi}$ is as follows:

\begin{equation}
\begin{aligned} 
    \mathcal{L}_{\text{critique}}(\phi) &=  \mathbb{E}_{(x,y,c) \sim \mathcal{D}_{\text{critique}}}\Big [\log{\pi_{\phi}(c|x,y)}\Big ], 
\end{aligned}
\end{equation}

In this way, we can obtain a critique model that provides step-level supervision and constructive feedback on reasoning paths for actor models. 

\paragraph{Training actor models with basic reasoning and refinement ability.}

We then train reasoning models in our two-player setting. The models are trained using the training sets of GSM8K and MATH, containing $7,473$ and $7,500$ samples, respectively. We denote the mixed response training set as \(\mathcal{ D}_{\text{reason}} = \{(x, y)\}_{j=1}^{|\mathcal{D}_{\text{reason}}|} \). Additionally, to equip the models with the ability to perform refinement tasks according to the critique feedback, we utilize GPT-4 to annotate $8$k refinement samples ( half of which are from MATH and the other half from GSM8K), denoted as \( \mathcal{D}_{\text{refine}} = \{(x, y, c, y')\}_{j=1}^{|\mathcal{D}_{\text{refine}}|} \), where \( y' \) represents the refined reasoning path generated based on the critique \( c \). Each refinement sample is verified to ensure the correctness of its final answer. The loss of training actor reasoning model $\pi_\theta$ is as follows:

\begin{equation}
\begin{aligned} 
    \mathcal{L}_{\text{actor}}(\theta) &=  \mathbb{E}_{(x,y) \sim \mathcal{D}_{\text{reason}}}\Big [\log{\pi_{\theta}(y|x)}\Big ]  + \beta \times \mathbb{E}_{(x,y,c,y') \sim \mathcal{D}_{\text{refine}}}\Big [\log{\pi_{\theta}(y'|x,y,c)}\Big ] , \label{eqn: actor loss}
\end{aligned}
\end{equation}

where $\beta$ is a hyper-parameter that balances the learning of reasoning and refining.

\subsection{Critique-based Supervision Improves Test-time Reasoning Performance} \label{sec: Motivating Insight: Critique-in-the-loop Improves Test-time Performance}

In this section, we investigate the impact of trained critique models in supporting the reasoning model at test-time (illustrated on the left of Figure \ref{fig: test-time-and-training-time-illustration}). Specifically, we examine their effectiveness in enhancing the actor’s reasoning performance, identify the types of problems where performance improvements are observed, and assess whether scaling up test-time computation further elevates the actor’s performance ceiling.

\subsubsection{Experimental Setups}
\paragraph{Backbone models.}
In our main experiments, we fine-tune the actor models using Llama3-8B-Base, following previous work \citep{DBLP:journals/corr/abs-2407-18219,DBLP:journals/corr/abs-2406-03816, DBLP:journals/corr/abs-2409-12917}. This model demonstrates non-trivial performance on mathematical reasoning tasks while leaving room for improvement, making it an ideal testbed for our study.
We fine-tune the critique models using the fine-tuned models Llama3-8B and Llama3-70B, which have the instruction-following ability to serve as our critique backbone. Note that most of our experiments are performed with the 8B model.

\paragraph{Evaluation metrics.}

In mathematical reasoning tasks, we primarily evaluate the accuracy, which measures whether a solution matches the ground truth with an oracle reward function. When critique models are not employed, we directly evaluate the accuracy of the actor's responses. In contrast, when critique models are used, we evaluate the accuracy of the actor's responses after refinement based on feedback provided by the critique model.

Additionally, to comprehensively assess a critique model, we evaluate its discriminability, i.e., the ability to determine whether a solution contains errors \citep{DBLP:journals/corr/abs-2206-05802}. We also evaluate its helpfulness, which means whether it can provide constructive feedback that enables the actor to correct erroneous responses.

\paragraph{Implementation details.}
The experiments are conducted on NVIDIA A100 GPUs and Ascend 910 processors. When fine-tuning the critique models and actor reasoning models, we set the learning rate to $2e-5$. During decoding, we set the model's temperature to $0$, which means the decoding process is done greedily. When we scale up inference-time computation, we set the temperature to $0.7$. 
We evaluate the accuracy of the actor models, and the discriminability and helpfulness of the critique models.

\subsubsection{Empirical Results and Findings}
\begin{table*}[t]
\centering
\caption{Test-time evaluation results of critique models on GSM8K and MATH. ``Acc.'' represents accuracy; ``Discrimin.'' refers to the accuracy of determining whether a reasoning path contains errors; ``Helpfulness'' indicates the ability of critique models to provide assistance for an incorrect reasoning path. The ``No Critic'' baseline represents the standalone performance of the actor reasoning model. Our 8B critique model outperforms GPT-3.5-Turbo, while the 70B critique model achieves performance close to the GPT-4 series models.}
\vspace{2pt}
\resizebox{0.99\textwidth}{!}{ 
\begin{tabular}{lccc|ccc}
\toprule
\multirow{2}{*}{\textbf{Critique Model}} & \multicolumn{3}{c}{\textbf{GSM8K}}              & \multicolumn{3}{c}{\textbf{MATH}}              \\
& \multirow{1}{*}{\textbf{Acc.}}  & \multirow{1}{*}{\textbf{Discrimin.}} & \multirow{1}{*}{\textbf{Helpfulness}} & \multirow{1}{*}{\textbf{Acc.}}  & \multirow{1}{*}{\textbf{Discrimin.}} & \multirow{1}{*}{\textbf{Helpfulness}} \\ 
\midrule
No Critic & $54.81$ & - & - & $17.22$ & - & - \\
GPT-3.5-Turbo & $58.38$ & $62.9\%$ & $13.3\%$ & $25.56$ & $51.3\%$ & $14.3\%$ \\
GPT-4-Turbo & $77.86$ & $91.6\%$ & $57.5\%$ & $36.00$ & $87.6\%$ & $26.2\%$ \\
GPT-4o & $79.52$ & $91.5\%$ & $59.7\%$ & $39.98$ & $85.4\%$ & $30.9\%$ \\
Critique Model-8B & $63.31$ & $79.4\%$ & $31.0\%$ & $24.26$ & $75.7\%$ & $16.2\%$ \\
Critique Model-70B & $76.88$ & $92.3\%$ & $55.3\%$ & $33.94$ & $82.3\%$ & $23.9\%$ \\
\bottomrule
\end{tabular}
}
% \vspace{-4mm}
\label{tab:test_time_performance}
\end{table*}

\paragraph{Critique models are highly effective at identifying the correctness of reasoning, offering constructive feedback for erroneous responses, and improving the overall accuracy of the actor.}

We compare our critique models with SOTA models used as critics, and the results are presented in Table \ref{tab:test_time_performance}. We observe that compared to current state-of-the-art (SOTA) models, our 8B critique model significantly outperforms GPT-3.5, while our Llama3-Critic-70B model achieves performance comparable to GPT-4 series models. 

Specifically, the reasoning path judgment accuracy of our 8B critique model reaches $79.37\%$ on GSM8K and $75.74\%$ on MATH, exceeding GPT-3.5-Turbo by $16.52$ and $24.46$ percentage points, respectively. Additionally, in terms of helpfulness, it outperforms GPT-3.5-Turbo by $17.70\%$ and $1.93\%$ on GSM8K and MATH, respectively. 
Moreover, our 70B critique model demonstrates even stronger performance. As to discriminability, it surpasses GPT-4-Turbo and GPT-4o on the GSM8K dataset and achieves results close to these SOTA models on MATH. Its correction accuracy on both datasets approaches that of GPT-4 series models, ultimately leading to comparable actor accuracy under its guidance.

\begin{figure*}[t]
    \includegraphics[width=0.99999\linewidth]{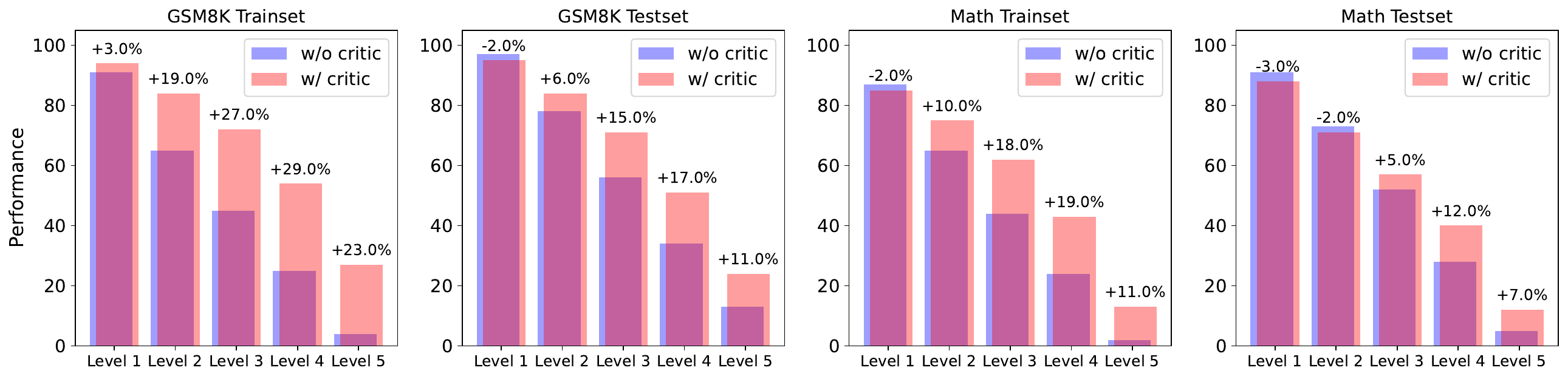}
    \centering
    \vspace{-0.3cm}
 	\caption{
    The impact of using critique models on performance across different difficulty levels. Critique models help the actor achieve better performance on more challenging queries.
    }\label{fig:difficulty_distribution}
  \vspace{-0.18cm}
\end{figure*} 

\paragraph{Critique models assist the actor in better handling challenging queries.} 
Next, we investigate the distribution of performance gains brought by the critique model across different difficulty levels. 
The process involves generating $100$ responses from the actor model for each query and categorizing the queries into 5 difficulty levels based on the number of correct responses associated with each query \cite{DBLP:conf/nips/HendrycksBKABTS21}.
The results are illustrated in Figure \ref{fig:difficulty_distribution}.
It is evident that on both the training and test sets of GSM8K and MATH, critique models provide minimal benefit for simpler queries, as the actor model can independently perform well in these cases. However, for more challenging problems, critique models offer significant support, resulting in overall improved performance. Furthermore, this phenomenon is even more pronounced in the training set, offering valuable insights for incorporating critique model supervision during training (Section \ref{sec: Critique-in-the-loop Self-Improvement for Better Reasoning Models}) \citep{DBLP:journals/corr/abs-2407-13690}.

\paragraph{Scaling up inference-time computation consistently improves reasoning performance.}

Recent studies have highlighted that scaling up inference-time computation can significantly enhance model performance \citep{DBLP:journals/corr/abs-2408-03314,DBLP:journals/corr/abs-2407-21787,DBLP:journals/corr/abs-2408-16737}. Here, we investigate whether incorporating critique models can further elevate the reasoning performance ceiling as test-time computation scales. 
A widely used technique employed in test-time computation scaling is majority voting \citep{DBLP:conf/iclr/0002WSLCNCZ23}, denoted as Maj@K, which measures whether the most frequent answer among $K$ parallel samples is correct. 
This metric reflects the model's consistency in generating high-quality responses across multiple samples, which is a critical aspect of interactive exploration and learning paradigms such as reinforcement learning and self-improvement.

As shown in Figure \ref{fig:MV_test_time_scaling}, without critique models, Maj@K performance improves with increased computation but quickly plateaus, even at higher levels of computation (e.g., Maj@2K, Maj@3K). In contrast, when critique models are utilized during test-time, performance surpasses the baseline by a significant margin under the same computation budget—showing a $12.4\%$ improvement on GSM8K and a $14.8\%$ improvement on MATH. These findings indicate that critique models effectively improve the exploration efficiency and quality of critique models, extending the performance ceiling when allocated more inference-time computation.

\section{Critique-in-the-loop Self-Improvement for Better Reasoning Models}
\label{sec: Critique-in-the-loop Self-Improvement for Better Reasoning Models}

Motivated by the test-time findings in Section \ref{sec: Motivating Insight: Critique-in-the-loop Improves Test-time Performance} that critique models significantly aid in solving challenging problems, and that they substantially raise the reasoning performance ceiling when scaling up computation, we integrate the critique-based supervision into the actor model's iterative exploration and learning process. We present a critique-in-the-loop self-improvement method, which scales up exploration computation on challenging queries and leads to the development of stronger reasoning models (illustrated in Figure \ref{fig: test-time-and-training-time-illustration}).

\subsection{Vanilla Self-Improvement Method}

Self-improvement is an exploration and learning method \citep{DBLP:conf/emnlp/0001GHW00023,DBLP:journals/corr/abs-2312-10003,DBLP:journals/corr/abs-2407-18219,DBLP:journals/corr/abs-2312-10003,tian2024toward,DBLP:conf/emnlp/DouY0CP24}. It iteratively leverages the actor reasoning model's correct responses to gradually enhance its problem-solving abilities. The process involves \( T \) iterations, where each iteration consists of two steps: exploration and learning.

In the \textbf{exploration} step of iteration \( t \), we sample \( N \) responses for each query \( x_j \in \mathcal{D}_{\text{reason}} \) from the previous model \( \pi_{\theta}^{t-1} \), i.e., \( \hat{y}_j = \pi_{\theta}^{t-1}(x_j) \). Each data point is then filtered using the reward function \( r(x, y) \), where only correct solutions are retained to form a new dataset \( \mathcal{D}^{t} = \{(x, y)\}_{j=1}^{|\mathcal{D}^{t}|} \).

In the \textbf{learning} step of iteration \( t \), the new dataset from the exploration step is used to fine-tune the actor reasoning model \( \pi_{\theta} \). 
To mitigate overfitting, we follow recent work \citep{DBLP:conf/nips/ZelikmanWMG22} and always fine-tune the original model \( \pi_{\theta}^0 \) instead of the model from the previous step, \( \pi_{\theta}^{t-1} \). 
The training loss is as in Equation \ref{eqn: actor loss} and we also include the original reasoning set $\mathcal{D}_{\text{reason}}$ and refinement set $\mathcal{D}_{\text{refine}}$ \citep{DBLP:journals/corr/abs-2308-08998}. After the improve step is performed, a new dataset of better quality samples can be created once again \citealp{DBLP:journals/corr/abs-2406-04151}.

\paragraph{Limitations of vanilla self-improvement.}

In self-improvement, the key challenge lies in identifying correct responses with high diversity for each query during the exploration step \cite{DBLP:conf/emnlp/0001GHW00023,tian2024toward}. 
However, previous studies have highlighted the problem known as the tail narrowing \citep{ding2024mitigating}. Specifically, models tend to over-sample solutions for simpler queries while under-sampling solutions for harder queries. 
This results in a training set for the next iteration that contains a large number of solutions for simple problems but lacks solutions for more challenging problems, introducing sampling bias. As iterations progress, this bias deepens, leading to a long-tail distribution where solutions for harder queries are almost entirely absent. This ultimately causes the model to reach a performance plateau or even degrade \citep{ding2024mitigating}.

\subsection{Critique-in-the-loop Self-improvement}

\begin{algorithm}[t]
\caption{Critique-in-the-loop Self-Improvement}
\label{Algorithm: Critique-in-the-loop Self-Improvement}
  \SetKwData{Left}{left}\SetKwData{This}{this}\SetKwData{Up}{up}
  \SetKwFunction{Union}{Union}\SetKwFunction{FindCompress}{FindCompress}
  \SetKwInOut{Input}{Input}\SetKwInOut{Output}{Output}
   \SetKwProg{myproc}{Procedure}{}{}
   \KwIn{Initialized actor reasoning model $\pi_\theta$, intialized critique model $\pi_{\phi}$, reasoning dataset $\mathcal{D}_\text{reason}$, refinement dataset $\mathcal{D}_\text{refine}$, critique dataset $\mathcal{D}_\text{critique}$, oracle reward function $r$, the iteration number for self-improvement $T$, the sampling number for exploration $N$, the sampling number for critique generation $L$.}
   \myproc{{{\textnormal{Fine-tune the critique model and the actor reasoning model}}}}{  
        Minimize the following loss objective to obtain critique model $\pi_{\phi}$: \\
        \ \ \ \ \ $\mathcal{L}_{\text{critique}}(\phi) =  \mathbb{E}_{(x,y,c) \sim \mathcal{D}_{\text{critique}}}\Big [\log{\pi_{\phi}(c|x,y)}\Big ]$;   \\
        Minimize the following loss objective to obtain actor model $\pi_{\theta}^{\text{base}}$:\\ 
        \ \ \ \ \ $\mathcal{L}_{\text{actor}}(\theta) =  \mathbb{E}_{(x,y) \sim \mathcal{D}_{\text{reason}}}\Big [\log{\pi_{\theta}(y|x)}\Big ]  + \beta \times \mathbb{E}_{(x,y,c,y') \sim \mathcal{D}_{\text{refine}}}\Big [\log{\pi_{\theta}(y'|x,y,c)}\Big ] $;   \\
    }
  \myproc{{{\textnormal{Exploration and Learning with Critique Supervision}}}}{  
    
   $\pi_{\theta}^0 \gets \pi_{\theta}^{\text{base}}$; \\
        \For{\ \textnormal{iteration} \  $t=1$ \textnormal{to} $T$ }{
            \myproc{{{\textcolor{violet!70!red}{\textnormal{Exploration Step}}}}}{ 
            $\mathcal{D}^{t} \gets \varnothing$; \\
            \textcolor{gray}{// Sample $N$ solutions from the reasoning model and collect correct responses.} \\
            \For{\ \textnormal{sample num} \  $n=1$ \textnormal{to} $N$ }{
                $\mathcal{D}^{t,n} = \{(x_i,y_i)| x_i \sim \mathcal{D}_{\text{reason}}, y_i \sim \pi_{\theta}^{{t-1}}(y|x_i)\}$; \\
                $\mathcal{D}^t \gets \mathcal{D}^t \cup \mathcal{D}^{t,n}$; \\
            }
            Apply $r$ to $\mathcal{D}^{t}$ to get correct responses $\mathcal{D}^{t}_{\text{correct}}$ and incorrect responses $\mathcal{D}^{t}_{\text{incorrect}}$; \\
            \textcolor{gray}{// Generate critique and refinement for incorrect responses.} \\
            $\mathcal{D}_{\text{critique}}^t, \mathcal{D}^{t}_{\text{refine}} \gets \varnothing$; \\
            \For{\ \textnormal{critique generation num} \  $l=1$ \textnormal{to} $L$ }{
            $\mathcal{D}_{\text{critique}}^{t,l} = \{(x_i,y_i,c_i)| x_i,y_i \sim \mathcal{D}^{t}_{\text{incorrect}}, c_i \sim \pi_{\phi}(c|x_i,y_i)\}$; \\
             $\mathcal{D}_{\text{critique}}^{t} =  \mathcal{D}_{\text{critique}}^{t} \cup  \mathcal{D}_{\text{critique}}^{t,l} $; \\
            $\mathcal{D}^{t,l}_{\text{refine}} = \{(x_i,y_i,c_i,y_i')| x_i,y_i,c_i \sim \mathcal{D}^{t,l}_{\text{critique}}, y_i' \sim \pi_{\theta}^{{t-1}}(y|x_i,y_i,c_i)\}$;\\
            $\mathcal{D}^{t}_{\text{refine}} \gets \mathcal{D}^{t}_{\text{refine}} \cup \mathcal{D}^{t,l}_{\text{refine}}$; \\
            }
            \textcolor{gray}{// Combine correct original solutions and correct refined solutions.} \\
            Apply $r$ to $\mathcal{D}^{t}_{\text{refine}}$ to get the correct refine set $\mathcal{D}^{t}_{\text{correct\_refine}}$; \\
            $\mathcal{D}_{\text{correct}}^t \gets \mathcal{D}_{\text{correct}}^t \cup \{(x_i,y_i')| x_i,y_i' \sim \mathcal{D}^{t}_{\text{correct\_refine}}\}$ ;\\
            } 
            \myproc{{{\textcolor{red!30!cyan}{\textnormal{Learning Step}}}}}{ 
            $\mathcal{D}_\text{train}^t = \mathcal{D}_{\text{reason}} \cup \mathcal{D}_{\text{correct}}^t $; \\
            Minimize the following loss objective to obtain $\pi_{\theta}^{t}$:
            \ \ \ \ \ \ $\mathcal{L}_{\text{actor}}(\theta) =  \mathbb{E}_{(x,y) \sim \mathcal{D}_{\text{train}}}\Big [\log{\pi_{\theta}(y|x)}\Big ]  + \beta \times \mathbb{E}_{(x,y,c,y') \sim \mathcal{D}_{\text{refine}}}\Big [\log{\pi_{\theta}(y'|x,y,c)}\Big ] $;   \\
            }

        }
    }
\end{algorithm}

\paragraph{Introduce critique models for high-coverage exploration. }
Motivated by our prior findings in Section \ref{sec: Motivating Insight: Critique-in-the-loop Improves Test-time Performance} that critique models enable actors to achieve greater performance gains on harder queries, we introduce critique models to the self-improvement process and propose a critique-in-the-loop self-improvement approach. 

This method is built upon self-improvement, and during the exploration step of iteration $t$, critique models $\pi_{\phi}$ are instructed to provide feedback on the responses of actor $\pi_{\theta}^{t}$, and the actor is then prompted to perform refinements accordingly. After that, correct refinements are then added to the training set. Since we can assume the availability of an oracle reward function for the training set, critiques are only applied to incorrect responses, while correct responses are directly included in the dataset. This design minimizes the risk of low-quality critiques negatively affecting originally correct responses. In this way, we increase the coverage of solutions for harder queries, and significantly reduce the tail-narrowing problem \citep{ding2024mitigating}.

\paragraph{Difficulty-aware computation allocation for exploration.}

Furthermore, building on our previous findings in Section \ref{sec: Motivating Insight: Critique-in-the-loop Improves Test-time Performance} that scaling inference-time computation can improve the efficiency and quality of exploration, we allocate more computation for exploration and critique to harder problems. This involves performing additional response generation, critique, and refinement to obtain high-quality and diverse solutions. 

In practice, we employ a simple difficulty-based computation allocation strategy, as it has proven sufficiently effective.\footnote{Note that more complex strategies, such as distinguishing difficulty based on the accuracy observed after multiple samples, are expected to yield even better results.} 
For incorrect initial responses, we classify them as difficult and allocate $L$ times of critique and refinement. For correct initial responses, they are considered simple and are directly added to the training set without further critique or refinement. This approach further mitigates the long-tail issue of self-improvement, enhances sampling quality, and improves the overall performance \citep{ding2024mitigating}.

We summarize the critique-in-the-loop self-improvement method in Algorithm \ref{Algorithm: Critique-in-the-loop Self-Improvement}.

\subsection{Experimental Results and Findings}

\paragraph{Implementation details.}

We set the self-improvement process to run for $3$ iterations, as in previous works \citep{progress_or_regress}, the model's performance tends to saturate after $3$ iterations of exploration and learning. During the exploration stage, we set the temperature to $0.7$ and the number of samples to $5$ or $10$. During the learning stage, we set the learning rate to $2e-5$ and the number of epochs to $1$.

\paragraph{Critique-in-the-loop self-improvement consistently improves reasoning performance.}
\begin{figure*}[t]
    \includegraphics[width=0.9\linewidth]{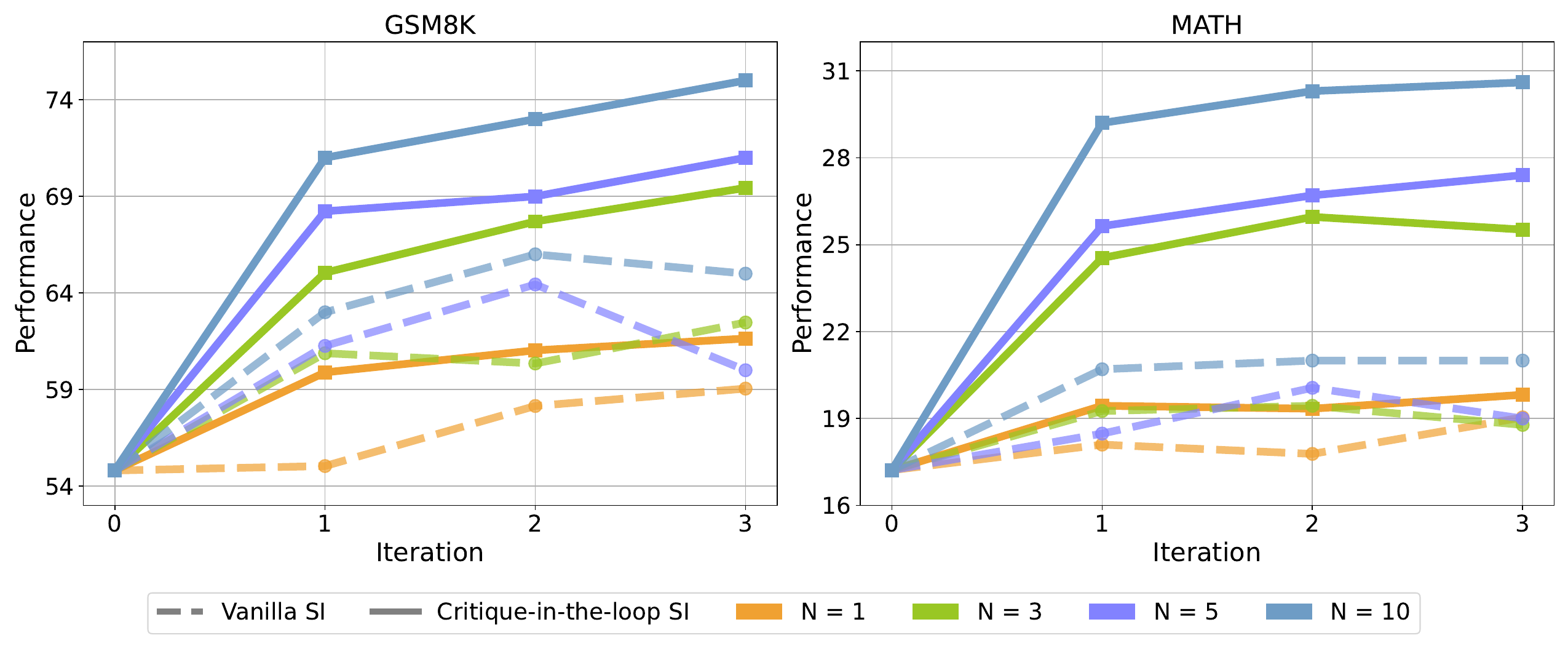}
    \centering
    \vspace{-1pt}
 	\caption{
    The evaluating results of critique-in-the-loop self-improvement. ``SI'' in the figure means self-improvement.
Compared to the vanilla self-improvement approach, our method achieves significant performance improvements, particularly at larger $N$ values.  }\label{fig:self_improve}
\end{figure*} 

The evaluating results of our method are shown in Figure \ref{fig:self_improve}. We can observe that: (1) Increasing the number of samples during exploration improves performance, with the performance upper bound rising accordingly, underscoring the benefits of enhanced exploration computation. (2) Our method consistently outperforms vanilla self-improvement with stable and significant performance gains, especially when the sample number $N$ is larger. For example, when $N=10$, our method achieves a performance advantage of $11.1\%$ on both GSM8K and MATH. (3) While the vanilla method initially shows performance improvements during self-improvement, it quickly reaches a bottleneck or even starts to decline, which may be attributed to the tail narrowing issue \cite{ding2024mitigating}. In contrast, our method demonstrates consistent improvement, with performance saturation occurring much later, indicating the effectiveness of our method.
\paragraph{Critique-in-the-loop self-improvement balances the solution distribution across difficulty levels, and enhances performance on challenging queries in the test set.}

\begin{figure*}[t]
    \includegraphics[width=0.45\linewidth]{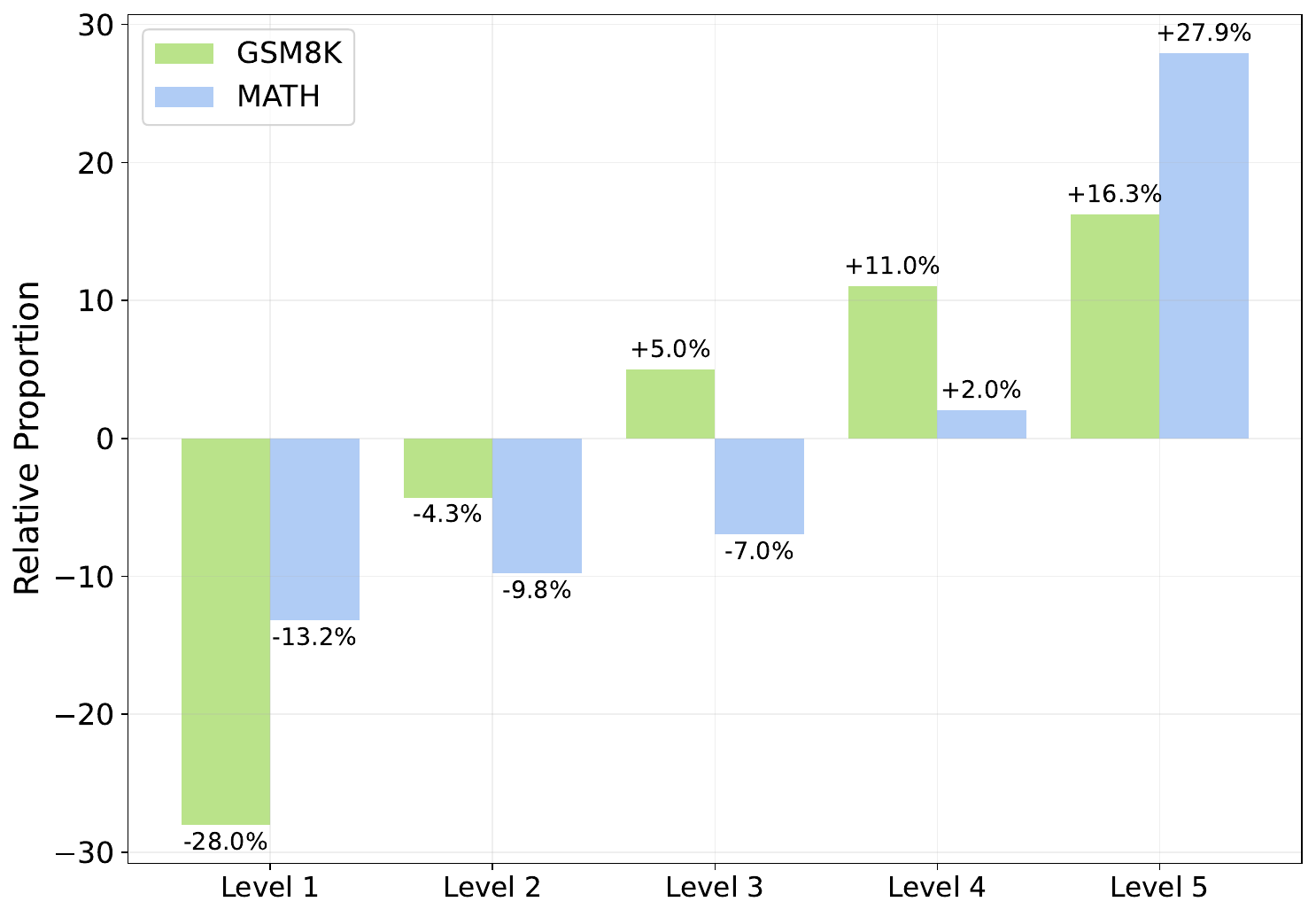}
    \centering
    \vspace{-0.3cm}
 	\caption{
    The difference in the proportion of training data across different difficulty levels obtained from the exploration steps of the critique-in-the-loop self-improvement method compared to the data obtained from the exploration steps of the vanilla self-improvement method. On both datasets, our method increases the proportion of difficult problems in the training set while reducing the proportion of simpler problems.
  }\label{fig:difficulty_propotion_difference}
  \vspace{-0.18cm}
\end{figure*}

\begin{figure*}[t]
    \includegraphics[width=0.8\linewidth]{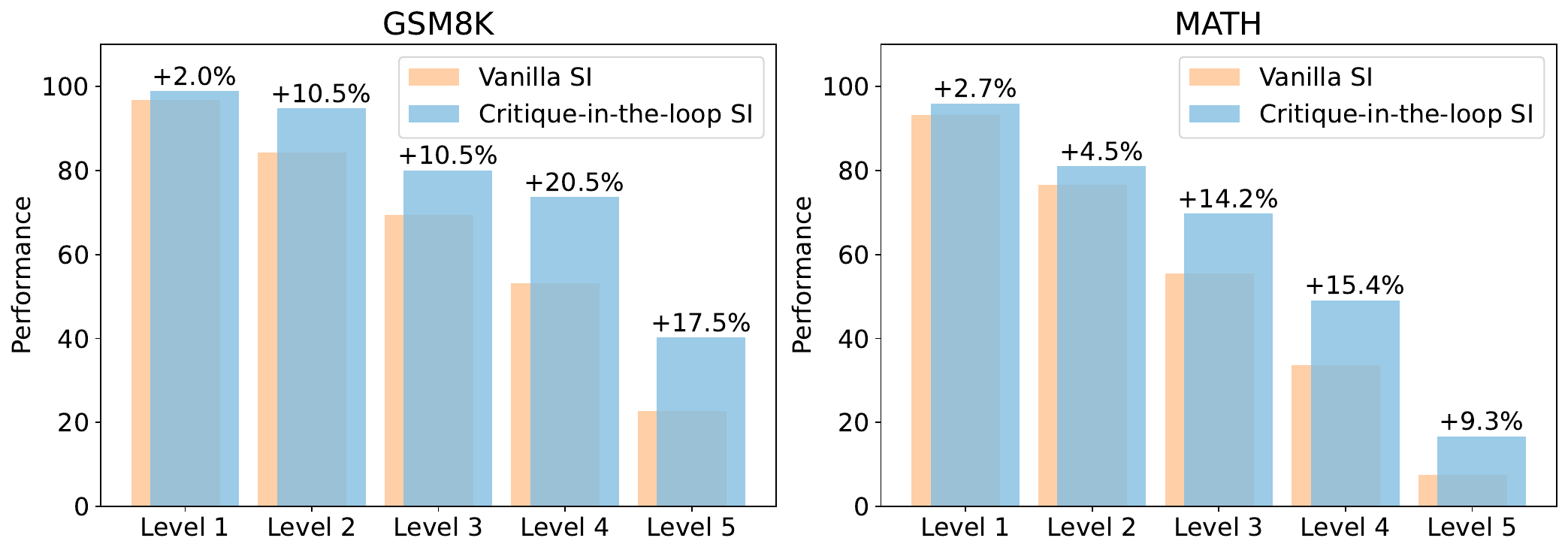}
    \centering
    \vspace{-0.3cm}
 	\caption{
    The performance differences between our method and vanilla self-improvement on test sets of varying difficulty. While our method slightly outperforms the vanilla approach on simpler problems, it achieves significantly greater improvements on harder problems.
}\label{fig:train_time_difficult_acc}
  \vspace{-0.18cm}
\end{figure*}

Since our motivation for introducing critique-based supervision into training is to improve the efficiency and quality of exploration, we examine the distribution of solutions sampled by our method compared to vanilla self-improvement. 
As shown in Figure \ref{fig:difficulty_propotion_difference}, we find that our approach samples a higher proportion of solutions for challenging queries during the exploration stage. 
This significantly balances the training data distribution for the learning stage, effectively mitigating the tail-narrowing issue. In Figure \ref{fig:train_time_difficult_acc}, we also present the model's performance on the test set across different difficulty levels, and we observe that our method performs significantly better than the vanilla approach on harder problems, further demonstrating the potential of our approach.

\paragraph{Combining test-time supervision with training-time supervisions yields more performance gains.}
\begin{table*}[t]
\centering
\caption{
Evaluation results of combining different training-time and test-time methods. During training, ``Self-Correction Fine-tuning'' refers to training a model with both reasoning and correction capabilities. For test-time methods, ``response only'' represents the actor model generating a response without additional correction or critique; ``w/ critique model'' indicates using a critique model at test-time to provide feedback, enabling the actor to perform refinement; and ``self-correction'' refers to the model generating a response and then performing correction by itself. The best performance is in \textbf{bold} and \underline{underlined}, while the second-best performance is \underline{underlined}.
From the results, we observe that both test-time and train-time critique supervision provide consistent improvements, and the combination of the two achieves the best performance.
}
\label{tab:combine_test_time_and_train_time_performance}
\vspace{3pt}
\resizebox{0.9999\textwidth}{!}{ 
\begin{tabular}{llcccccc}
\toprule
\multirow{2}{*}{\textbf{Training-time}} &\multirow{2}{*}{\textbf{Test-time}} & \multicolumn{3}{c}{\textbf{GSM8K}} & \multicolumn{3}{c}{\textbf{MATH}}  \\
% \cline{3-8}
 &  & \multicolumn{1}{c}{\textbf{Acc.}}  & \multicolumn{1}{c}{\textbf{Pass@5}} & \multicolumn{1}{c}{\textbf{MV@5}} & \multicolumn{1}{c}{\textbf{Acc}}  & \multicolumn{1}{c}{\textbf{Pass@5}} & \multicolumn{1}{c}{\textbf{MV@5}} \\ 
\midrule
\multirow{2}{*}{Supervised Fine-tuning}  & response only & $54.8$ & $75.2$ & $54.5$ & $17.2$ & $35.0$ & $15.6$ \\
& w/ critique model & $63.3$ & $87.6$ & $75.4$ & $24.3$ & $47.4$ & $30.7$ \\
\multirow{2}{*}{Self-Correction Fine-tuning} & response only & $54.2$ & $73.1$ & $53.4$ & $18.1$ & $32.4$ & $16.6$ \\
 & self-correction & $60.1$ & $81.5$ & $67.2$ & $24.2$ & $41.7$ & $26.1$ \\
\multirow{2}{*}{Vanilla Self-Improve} & response only & $64.6$ & $83.4$ & $70.6$ & $20.2$ & $38.5$ & $23.0$ \\
  & w/ critique model & $70.2$ & $\underline{90.8}$ & $78.2$ & $27.0$ & $48.8$ & $31.4$ \\
\multirow{2}{*}{Critique-in-the-loop Self-Improve}   & response only & $\underline{75.5}$ & $89.1$ & $\underline{80.1}$ & $\underline{31.3}$ & $\underline{51.0}$ & $\underline{35.1}$ \\ 
  & w/ critique model & $\underline{\textbf{75.8}}$ & $\underline{\textbf{91.8}}$ & $\underline{\textbf{82.8}}$ & $\underline{\textbf{31.4}}$ & $\underline{\textbf{53.1}}$ & $\underline{\textbf{36.8}}$ \\ 
 
\bottomrule
\end{tabular}
}
% \vspace{-4mm}
\end{table*}

Previously, we have evaluated the impact of incorporating critique model supervision at training and test-time, separately. Here, we combine them and evaluate the performance. Additionally, we include a self-correction baseline where the model refines its reasoning by itself. The training data for this baseline consists of original reasoning datasets (GSM8K and MATH) and correction data derived from our refinement data by removing critique elements, and reformatted into (query, original response, new response) triplets.

Evaluation results shown in Table \ref{tab:combine_test_time_and_train_time_performance} reveal that: (1) 
Integrating critique models during test-time consistently enhances performance under identical training conditions, particularly when critique supervision is not used during training. For example, applying critique models at test-time increases the MV@5 performance of SFT on GSM8K and MATH by $10.9$ and $15.1$ points, respectively. (2) When critique models are used during training, the additional benefit of test-time critique supervision becomes marginal, suggesting successful ``distillation'' of critique models into the actor during training. (3) The self-correction baseline underperforms compared to utilizing separate critique models, aligning findings in prior work that models struggle to accurately evaluate and refine their outputs without external feedback \citep{DBLP:conf/iclr/0009CMZYSZ24,DBLP:conf/acl/XuZZP0024}. Moreover, training a single model to handle both reasoning and correction capabilities may introduce conflicts, leading to performance degradation \citep{DBLP:conf/iclr/0009CMZYSZ24}. 
(4) Compared to the traditional strategy of vanilla self-improvement + response-only, which increases computation during training, the approach of supervised fine-tuning + test-time critique supervision reduces training computation while increasing test-time computation and achieves better performance, particularly on the more challenging MATH dataset. This aligns with prior work highlighting the benefits of enhancing test-time computation \citep{DBLP:journals/corr/abs-2408-03314,DBLP:journals/corr/abs-2407-21787,DBLP:journals/corr/abs-2408-16737}.

\paragraph{Ablation study.}

\begin{wraptable}[13]{r}{0.4\textwidth}
    \begin{minipage}{0.4\textwidth}
    \centering
    \vspace{-19pt}
\caption{Ablation study of critique-in-the-loop self-improvement.}
\resizebox{0.95\linewidth}{!}{
\begin{tabular}{lcc}
\toprule
\multirow{1}{*}{\textbf{Method}} & \multicolumn{1}{c}{\textbf{GSM8K}}              & \multicolumn{1}{c}{\textbf{MATH}}              \\
\midrule
\textbf{N=3}\\
\ \ \ \ Ours & $\underline{\textbf{69.1}}$   & $\underline{\textbf{25.5}}$   \\
\ \ \ \ w/o Difficulty Aware & $65.8$   & $23.9$   \\
\ \ \ \ More Refinement & $68.8$   & $25.5$   \\
\textbf{N=5}\\
\ \ \ \ Ours & $\underline{\textbf{72.3}}$   & $\underline{\textbf{28.0}}$   \\
\ \ \ \ w/o Difficulty Aware & $69.4$   & $26.6$   \\
\ \ \ \ More Refinement & $71.2$   & $27.7$   \\
\textbf{N=10}\\
\ \ \ \ Ours & $\underline{\textbf{75.4}}$   & $31.3$   \\
\ \ \ \ w/o Difficulty Aware & $70.1$   & $27.9$   \\
\ \ \ \ More Refinement & $73.8$   & $\underline{\textbf{31.6}}$   \\
\bottomrule
\end{tabular}
}
% \vspace{-4mm}
\label{tab:difficulty_aware_comparsion}
    \end{minipage}
\end{wraptable}

We evaluate the impact of using the strategy of difficulty-aware computation allocation for exploration, as well as the performance differences when allocating more computation to critique generation v.s. refinement generation. The experimental results are presented in Table \ref{tab:difficulty_aware_comparsion}, and we observe that:
(1) If the difficulty-aware allocation of exploration computation is removed, performance drops significantly. (2) If more allocation is used to generate multiple refinements for the same critique instead of generating diverse critiques, performance also decreases. Therefore, both difficulty-aware computation allocation and diversified critiques are crucial components for achieving the final performance.

\section{Discussion and Analysis}
\label{sec:Discussion and Analysis}
\subsection{Scaling Properties of Critique Models}
\begin{figure*}[t]
    \includegraphics[width=0.9\linewidth]{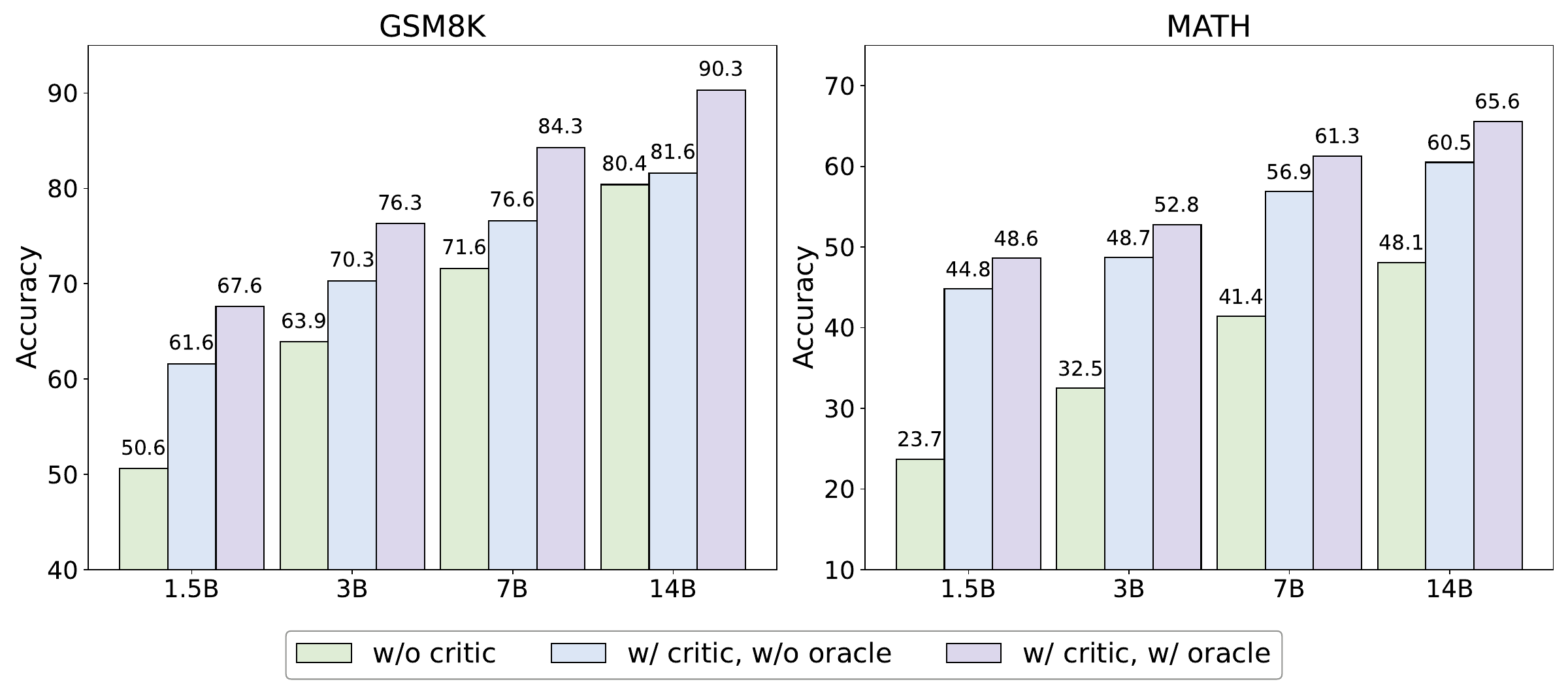}
    \centering
    \vspace{-1pt}
 	\caption{
   Scaling properties of critique models. The experiments are conducted on the Qwen-2.5 series models of varying sizes. We evaluate the impact of 3B-scale critique models on actors of different sizes. In this context, ``w/o critic'' refers to the direct outputs of the actor model, while ``w/ critic'' represents outputs refined by the actor with feedback from the critique model. In addition, ``oracle'' indicates whether an oracle reward function is used to assist the critique model in making judgments. With an oracle reward function, only incorrect responses are passed to the critique model; otherwise, all responses are evaluated by the critique model. The trained 3B critique model provides effective supervision for actors of various sizes.  }\label{fig:scale_properties}
\end{figure*} 

As in previous work \citep{DBLP:journals/corr/abs-2110-14168}, we study the scaling properties of critique models, trying to investigate whether they can supervise models of different scales, particularly those larger and stronger than themselves. In this study, we conduct experiments using the Qwen-2.5 series of models \citep{qwen2.5}, which span a wide range of scales (1.5B, 3B, 7B, and 14B). We train a critique model of 3B scale and use it to supervise trained actor reasoning models of all sizes. Other experimental settings are consistent with Section \ref{sec: Motivating Insight: Critique-in-the-loop Improves Test-time Performance}.

The evaluating results are shown in Figure \ref{fig:scale_properties}. In the figure, ``oracle'' indicates whether we have an oracle reward function to assist the critique model in making judgments. With an oracle reward function, only incorrect responses are passed to the critique model; otherwise, all responses are passed to the critique model. From the results, we can observe that: (1) Regardless of scale, the 3B critique model can provide effective supervision, indicating that smaller critique models can help supervise larger actors to a certain extent. (2) With the oracle reward function, the critique model does not need to perform discriminative tasks and only needs to provide useful feedback, resulting in greater performance improvements. (3) As the model scale increases, the performance improvement provided by the critique model on simpler datasets like GSM8K becomes marginal. However, on the more challenging MATH, the critique model continues to deliver significant performance gains even for the largest model.

\subsection{How do Critique Models improve Majority Voting?}
\begin{figure*}[t]
    \includegraphics[width=0.999\linewidth]{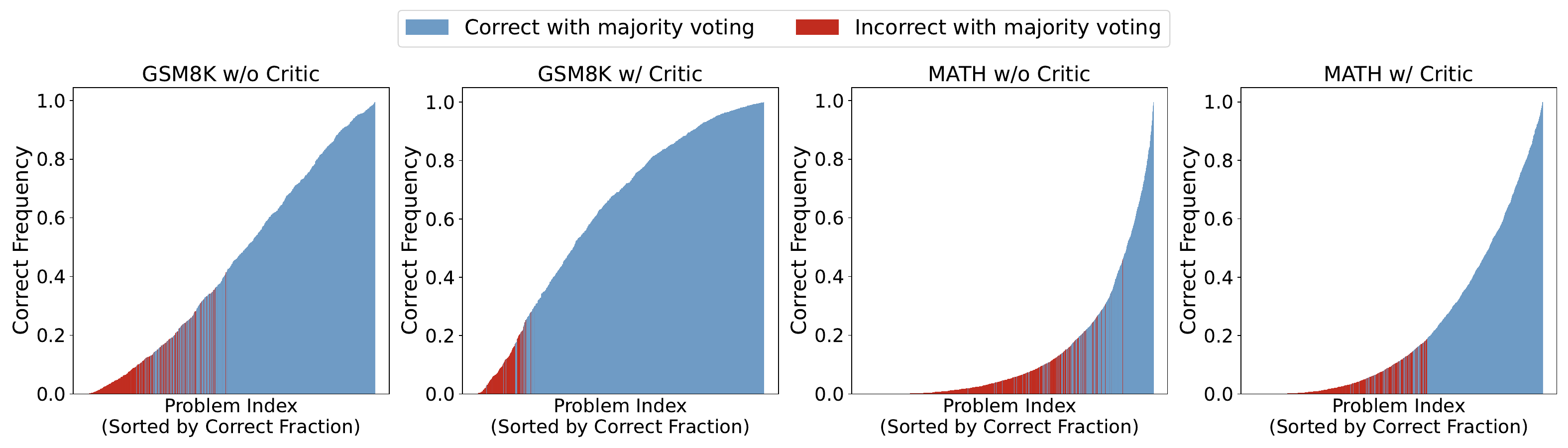}
    \centering
    % \vspace{-1pt}
 	\caption{
    Bar charts showing the fraction of correct samples (out of 1,000 samples) for each query in the test sets of GSM8K and MATH. Each bar represents a query, and the height corresponds to the fraction of samples that reach the correct answer. The bars are sorted by the correct fraction. Blue bars indicate that majority voting selected the correct answer, while red bars indicate that it did not. Note that ``w/o Critic'' means that no critique model is involved during the test process, while ``w/ Critic'' indicates that a critique model is used. Additionally, we do not assume the availability of an oracle reward function to assist critique models in making judgments. Critique models increase the fraction of correct answers and mitigate a failure mode where the correct answer appears frequently (e.g., $40\%$) but a specific incorrect answer occurs even more often (e.g., $45\%$), leading to majority voting selecting the wrong answer.
  }\label{fig:MV1000}
\end{figure*}

Majority voting is one of the most commonly used techniques for scaling test-time computation. Following previous work \citep{DBLP:journals/corr/abs-2407-21787}, we study the relationship between the correct frequency of multiple samples and the performance of majority voting, while also examining the impact of critique models. Specifically, consistent with the settings in \ref{sec: Motivating Insight: Critique-in-the-loop Improves Test-time Performance}, we use an actor reasoning model and a critique model trained with supervised fine-tuning. For each query, we sample $1,000$ responses in parallel. The experimental results are shown in Figure \ref{fig:MV1000}. 

We observe that critique models improve both the overall correct frequency and the performance of majority voting. Delving deeper, we can find a significant failure mode when critique models are not used, where the correct answer appears with a relatively high frequency (e.g., $40\%$), but a specific incorrect answer dominates in frequency (e.g., $45\%$), causing majority voting to select the incorrect result. By incorporating critique models, this failure mode is effectively addressed through critique and refinement. Specifically, the discriminative ability of critique models helps suppress the occurrence of high-frequency incorrect answers, while the feedback provided by these models increases the relative frequency of correct answers. Together, these mechanisms contribute to a substantial improvement in the performance of majority voting.

\subsection{Should test-time computation be scaled sequentially or in parallel?}
\begin{figure*}[t]
    \includegraphics[width=0.9\linewidth]{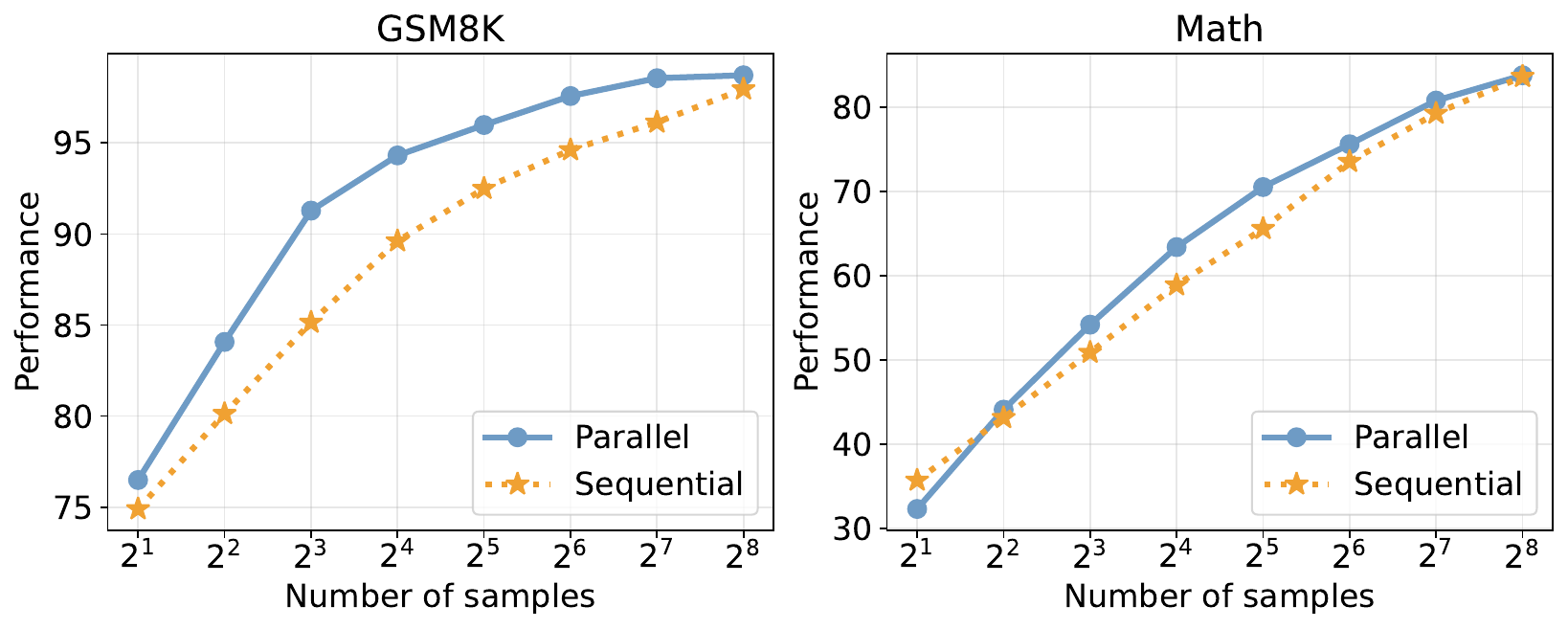}
    \centering
    \vspace{-0.3cm}
 	\caption{
    Pass@K performance of different strategies for scaling up test-time computation. Generating (response, critique, refinement) in parallel outperforms generating critiques and refinements sequentially.
  }\label{fig:test_time_pass}
  \vspace{-0.18cm}
\end{figure*}

\begin{figure*}[t]
    \includegraphics[width=0.9\linewidth]{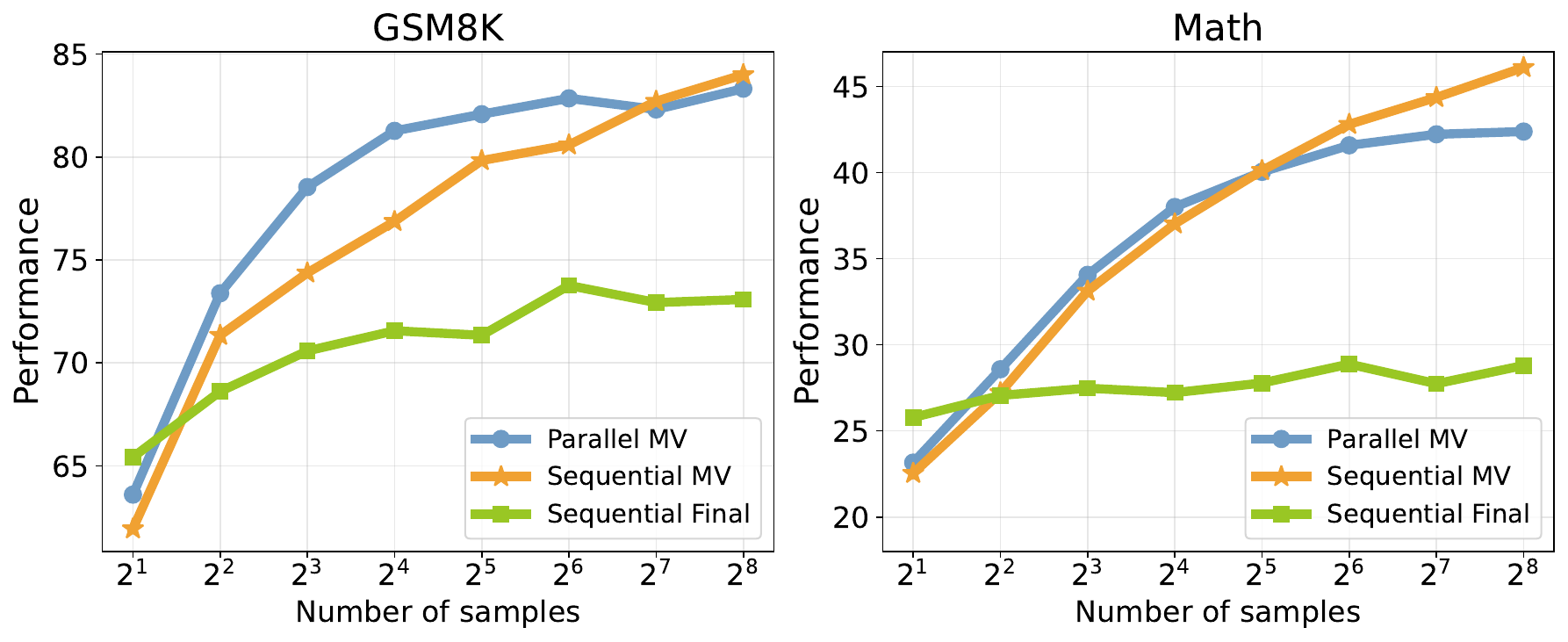}
    \centering
    \vspace{-0.3cm}
 	\caption{
    Evaluation results of different answer selection techniques when scaling up test-time computation. ``Parallel MV'' refers to generating \( K \) (response, critique, refinement) triplets in parallel and selecting the most frequent answer; ``Sequential MV'' refers to selecting the most frequent answer from all the linearly generated responses and refinements; ``Sequential Final'' selects the final answer after the \( K \)-th refinement.
  }\label{fig:test_time_mv}
  \vspace{-0.18cm}
\end{figure*} 

In the two-player paradigm, test-time computation can be scaled either in parallel by sampling multiple (response, critique, refinement) triplets \citep{DBLP:journals/corr/abs-2206-05802,DBLP:conf/iclr/MiaoTR24,DBLP:journals/corr/abs-2408-03314}, or sequentially by generating critiques and refinements iteratively after an initial response \citep{DBLP:conf/nips/ShinnCGNY23,DBLP:journals/corr/abs-2407-18219,DBLP:journals/corr/abs-2408-03314}. Here, we explore the performance of these two approaches. Notably, in our implementation of the sequential approach, to avoid potential issues of context window length limits, we use the following strategy: given a query \( x \), the actor first generates a response \( y_0 \). For the \( i \)-th critique task (\( i > 0 \)), only the query and the \( (i-1) \)-th response or refinement are provided. Similarly, for the \( i \)-th refinement task (\( i > 0 \)), only the query, the \( (i-1) \)-th response or refinement, and the \( i \)-th critique are provided.

Figure \ref{fig:test_time_pass} illustrates the Pass@K performance trends. Pass@K is a metric that measures whether at least one correct answer exists among $K$ samples. As computation increases, Pass@K performance improves, though the gains become marginal progressively. The performance of sequential computation scaling is slightly worse than that of parallel computation scaling. This may stem from the fact that in the parallel approach, the actor has more opportunities to generate original solutions directly from the queries, leading to greater sampling diversity and a higher chance of obtaining at least one correct answer. In contrast, the sequential approach allows the actor only one opportunity for original reasoning, with subsequent revisions relying on critical feedback, which may reduce diversity and limit the chances of finding correct solutions.

In Figure \ref{fig:test_time_mv}, we present the performance trends of three test-time techniques as computation increases: parallel majority voting, sequential majority voting, and sequential final. Here, sequential majority voting refers to selecting the most frequent answer from all the linearly generated responses and refinements, while sequential final selects the final answer after the \( K \)-th refinement. From the experimental results, we observe that: (1) Overall, compared to majority voting, selecting the final answer of a sequence of revisions performs weaker. This may be because the sequential critique and refinement process occasionally modifies a previously correct answer, resulting in an incorrect final result. (2) For smaller \( K \), parallel majority voting outperforms sequential majority voting. However, as computation scales up, sequential majority voting surpasses its parallel counterpart, especially on the more challenging MATH dataset. This indicates a trade-off between the parallel and sequential approaches \citep{DBLP:journals/corr/abs-2408-03314}, providing inspiration for our critique-in-the-loop self-improvement. Specifically, depending on the computation budget, the exploration strategy can be adapted to balance solution quality and diversity, ultimately leading to stronger actor reasoning models. We leave this study to future work.

\section{A Step Further: Training Step-level Self-Talk Reasoning Models via Critique Data}
\label{sec:A Step Further: Training Step-level Self-Talk Reasoning Models via Critique Data}

\paragraph{Motivation and method.} In this work, we focus on the two-player paradigm, leveraging critique models to provide step-level supervision and feedback for actor models. Recently, OpenAI's o1 model \citep{openai2024learning} has pushed the boundaries of large reasoning models' capabilities. With its self-talk output format, it can autonomously plan, reflect, critique, correct, backtrack, and more during the thinking process, marked by phrases such as ``wait'' and ``alternatively''. Therefore, we investigate whether it is possible to construct self-talk data with step-level critique supervision, and propose the preliminary self-talk-via-critique method. Specifically, it has three main steps:

\begin{enumerate}
    \item \textbf{Construct an initial thinking chain that has step-level reflection.} Given a query and a reasoning path, we first use AutoMathCritique to generate critique data. Feedback on each reasoning step provided in the critique is then inserted into the reasoning path, constructing a thinking chain that includes step-level reflections. At this stage, the thought process may lack smoothness but achieve an initial structure.
     \item \textbf{Iterative refine and critique the thinking chain.} For reasoning paths without errors, they are directly passed to the next stage. For paths containing errors, actors perform refinement from the first identified erroneous step, continuing the reasoning from that point onward. Starting from the first refined step, we utilize the critique model to re-critique the partial reasoning process—spanning from the refined step to the final step. Step-level feedback from this critique is again integrated into the thought process. If the critique model identifies new errors in the refined reasoning steps, the process is repeated iteratively. The reasoning path is continuously optimized until all errors are resolved by reflection and refinement. Only then is the thinking chain passed to the next stage.
    \item \textbf{Smooth the thinking chain into self-talk data.}
    Next, we prompt the LLMs to smooth out the previously rigid thinking chain. This involves adding transitional phrases and connectors to make the reasoning and reflection flow more naturally. Finally, we verify the correctness of the final answer, ensuring that only accurate data is stored.
\end{enumerate}

\begin{figure*}[t]
    \includegraphics[width=0.99999\linewidth]{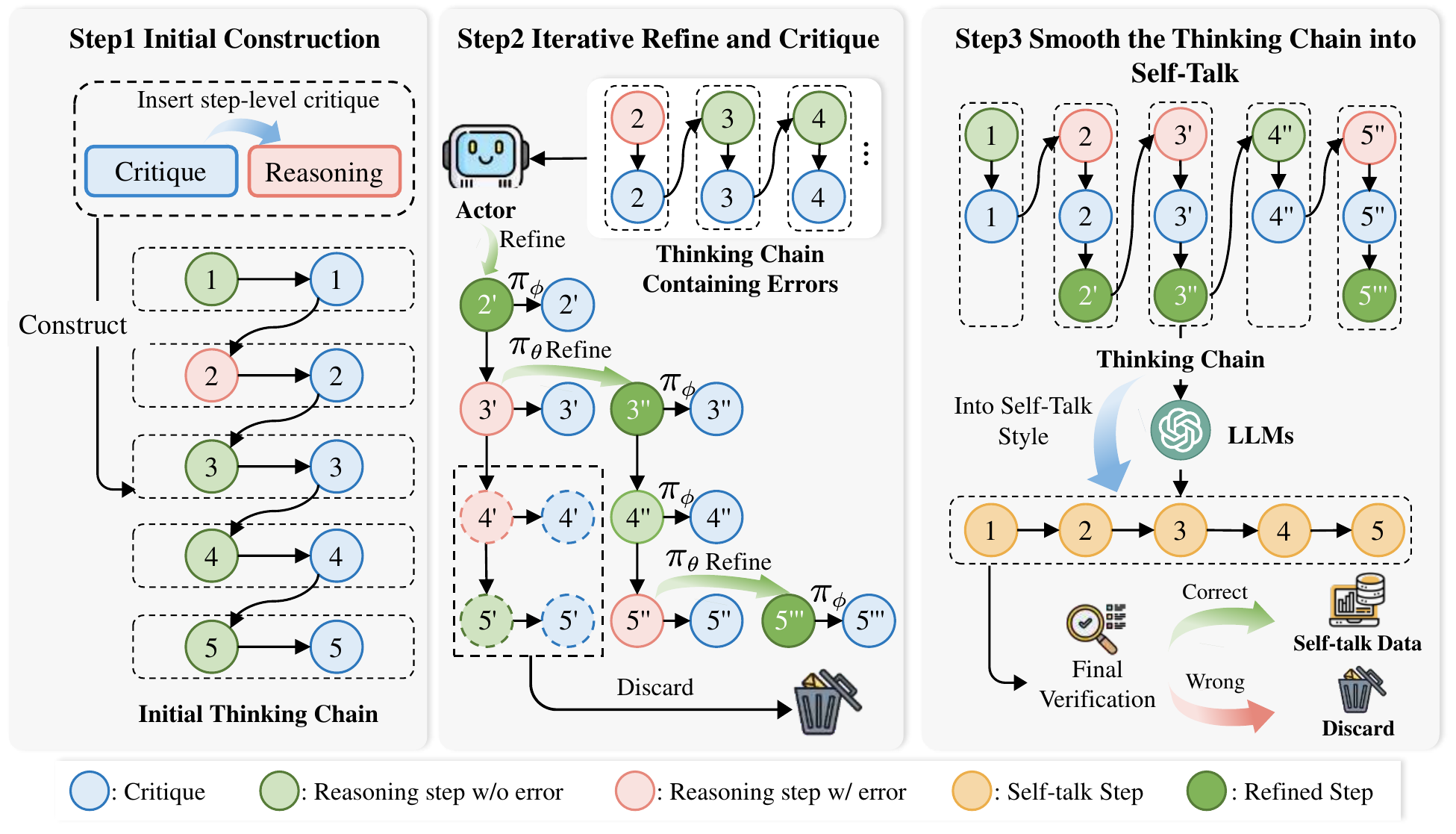}
    \centering
 	\caption{The overview of our self-talk-via-critique method to synthesize self-talk data. $\pi_{\theta}$ means the actor reasoning model, while $\pi_{\phi}$ means the critique model.}\label{fig:self-talk_process}
    \vspace{-8pt}
\end{figure*}

\begin{figure*}[t]
    \includegraphics[width=0.9\linewidth]{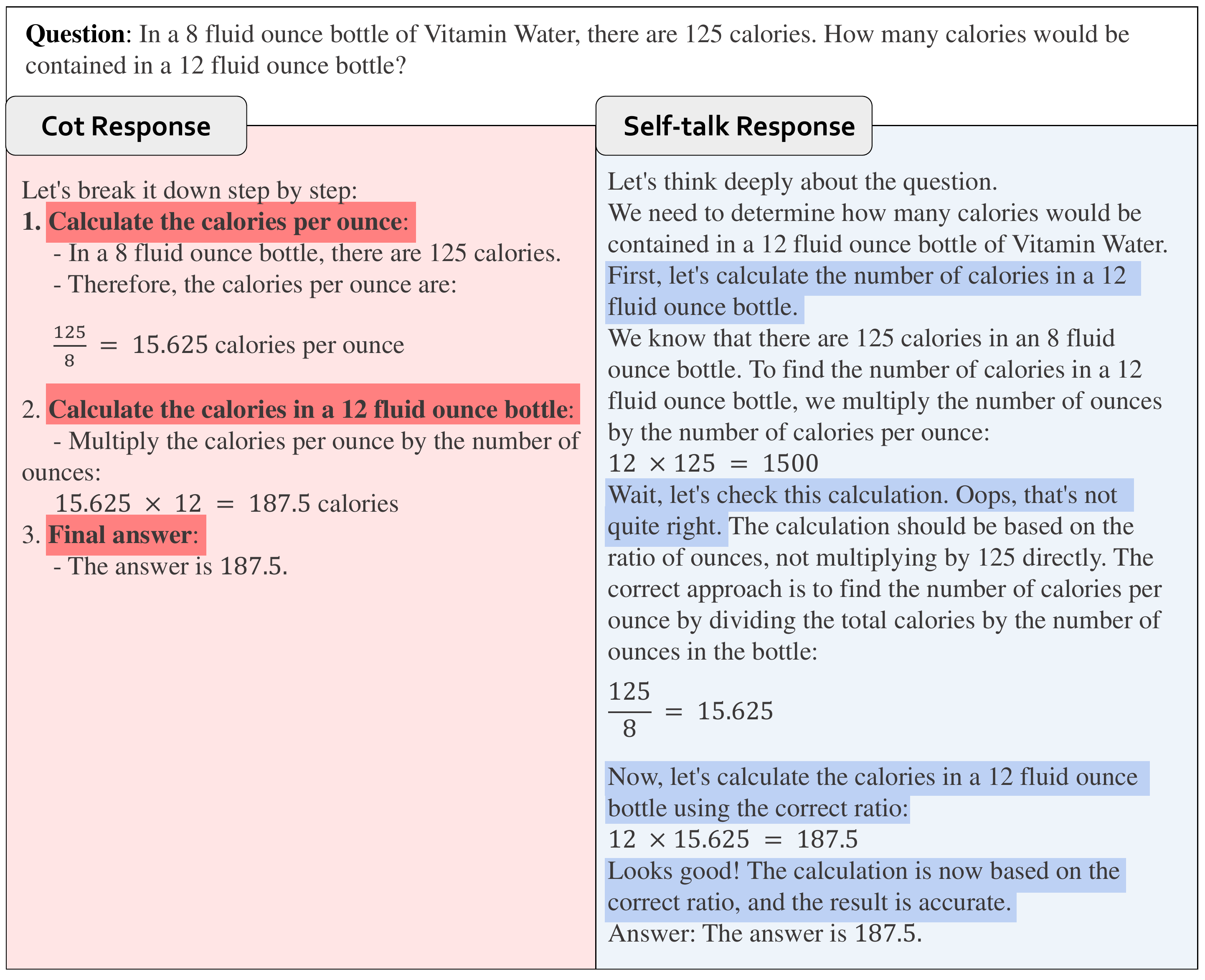}
    \centering
    % \vspace{-0.3cm}
 	\caption{ An example of the constructed step-leval self-talk data.
  }\label{fig:case_study_2}
  % \vspace{-0.18cm}
\end{figure*}

An overview of the method is shown in Figure \ref{fig:self-talk_process}. An illustrating example of the resulting self-talk data can be found in Figure \ref{fig:case_study_2}.

\paragraph{Evaluation and findings.}

Based on this approach, we construct a dataset of $4$k self-talk examples from the MATH training set and fine-tune the model for evaluation. As in the previous section, we used the Llama3-8B base model as the backbone for our experiments. We compared our method with the self-correction baseline and the baseline of SFT with test-time critique supervision. These two baselines fall under the one-player and two-player settings, respectively. Note that for the two baselines, we only used the MATH dataset for training, without using GSM8K data. The experimental results are shown in Table \ref{tab:self-talk}.
We observe that, in the one-player setting, the step-level self-talk approach outperforms trajectory-level self-correction by a significant margin, demonstrating its potential. However, it still lags behind the two-player setting, indicating that this direction requires further exploration, which we leave to future work.

\begin{table*}[t]
\centering
\caption{Evaluation results of self-talk-via-critique.}
\label{tab:self-talk}
\vspace{2pt}
\resizebox{0.8\textwidth}{!}{ 
\begin{tabular}{lccccc}
\toprule
\multirow{1}{*}{\textbf{Method}} & \multirow{1}{*}{\textbf{Acc}}  & \multirow{1}{*}{\textbf{Pass@5}} & \multirow{1}{*}{\textbf{MV@5}} & \multirow{1}{*}{\textbf{Pass@10}}  & \multirow{1}{*}{\textbf{MV@10}}  \\ 
\midrule
\textbf{One-player Setting}\\
\ \ \ \ Trajectory-level Self-Correction & $21.08$ & $38.12$ & $20.56$ & $46.64$ & $30.90$ \\
\ \ \ \ Step-level Self-talk & $24.90$ & $45.48$ & $28.44$ & $54.48$ & $31.28$ \\
\textbf{Two-player Setting} \\
\ \ \ \ SFT w/ test-time critic & $27.22$ & $49.16$ & $32.24$ & $58.86$ & $35.92$ \\
\bottomrule
\end{tabular}
}
% \vspace{-4mm}
\label{tab:selftalk_performance}
\end{table*}

\section{Related Work}
\paragraph{Training LLMs for reasoning through exploration and learning.}

Multi-step reasoning, such as mathematical reasoning and logical reasoning, is a challenging task for large language models (LLMs). Researchers have proposed prompting methods represented by Chain-of-Thought (CoT) to enable LLMs to think and reason step by step like humans, and then produce answers based on the reasoning process, significantly improving the model's reasoning performance \citep{DBLP:conf/nips/Wei0SBIXCLZ22, DBLP:conf/iclr/0002WSLCNCZ23}. 
To enhance the reasoning ability of models, previous work has focused on collecting large amounts of expert-labeled reasoning trajectories, allowing models to mimic step-by-step reasoning \citep{DBLP:conf/emnlp/MinLHALHZ22}. However, these methods are often difficult to scale up, as annotation is highly expensive, especially for very challenging and complex problems \citep{DBLP:journals/csur/Song0CMS23}.

Another category of methods, exploration and learning, seeks to address this issue by using model-generated data to train the model itself. Specifically, given a query, the model generates its own reasoning paths, and external supervision signals are used to filter out high-quality solutions, which are then used to train the model \citep{DBLP:conf/nips/ZelikmanWMG22,DBLP:conf/emnlp/0001GHW00023,DBLP:journals/corr/abs-2307-09288,DBLP:conf/acl/WangKMLSKH23,DBLP:journals/corr/abs-2312-06585}. This approach, also known as self-improvement or rejection sampling, often encounters the tail-narrowing problem, which can lead to performance bottlenecks \citep{DBLP:journals/corr/abs-2305-17493, DBLP:conf/iclr/AlemohammadCLHB24, ding2024mitigating}. Some researchers have proposed reinforcement learning-based approaches, where reward models are trained or oracle reward functions are used to provide supervision signals, enabling the model to explore and learn, thereby significantly improving reasoning performance \citep{DBLP:conf/iclr/LightmanKBEBLLS24,DBLP:journals/corr/abs-2308-09583, DBLP:conf/emnlp/ChenZWJHYZZXGZ024,DBLP:journals/corr/abs-2402-05808, DBLP:conf/acl/WangLSXDLCWS24}. However, reinforcement learning typically converges slowly, is costly, and poses challenges in providing reliable and dense process-level supervision signals.

In this work, we fine-tune critique models to provide reliable step-level supervision signals and helpful feedback during both training and test time. This approach improves sampling efficiency and quality, ultimately enhancing the actor's reasoning performance.

\paragraph{Developing models for critique, reflection and correction.}
Developing models with the ability to critique, reflect, and correct is an important way for scalable supervision and has been explored in various domains, such as summarization, mathematical reasoning, sequential decision-making, and coding \citep{DBLP:conf/nips/MadaanTGHGW0DPY23, DBLP:journals/corr/abs-2303-16749,DBLP:conf/iclr/WelleckLWBSK023,DBLP:conf/emnlp/XiJZZGLGZH23,DBLP:conf/eacl/PaulIPBBWF24}. 
Most previous work has used prompting techniques to have models generate critical comments or corrections about their own outputs \cite{bai2022constitutional,DBLP:conf/nips/MadaanTGHGW0DPY23,DBLP:conf/acl/DhuliawalaKXRLC24}.
However, these methods typically perform poorly without assuming an oracle reward function. This is because they struggle to assess their outputs correctly in the absence of external feedback, especially when the problems are more challenging \cite{DBLP:conf/iclr/0009CMZYSZ24,DBLP:conf/acl/XuZZP0024}. As a result, many fine-tuning or RL approaches are proposed to trained models to develop the capabilities \cite{DBLP:journals/corr/abs-2308-04592,DBLP:journals/corr/abs-2406-14024,DBLP:conf/iclr/ZhouWLSLQLJSZ024,DBLP:conf/icml/HavrillaRNDZHR24,DBLP:journals/corr/abs-2408-15240,DBLP:journals/corr/abs-2408-11791}.
The former often requires extensive human annotation, while the latter necessitates engineered and cumbersome reward designs. 
Another line of work leverages self-training ways to develop self-correction capabilities, e.g., ~\cite{DBLP:conf/iclr/WelleckLWBSK023,zheng2024criticcotboostingreasoningabilities,DBLP:journals/corr/abs-2409-12917,DBLP:journals/corr/abs-2407-18219}.
In contrast to these methods, in this paper, we delve into a two-player framework (e.g., \cite{DBLP:conf/acl/AkyurekAKCWT23,DBLP:conf/iclr/YaoHNLFXNC0AXMW24,DBLP:conf/acl/ZhouDL24}), distinguishing the roles of the critique model and the actor model. We propose a scalable data synthesis framework, AutoMathCritique, to generate critique data and train critique models. The trained critique models can provide supervision to yield stable performance gains for the actor model, both at test time and training time.

\paragraph{Scaling test-time computation for LLM Reasoning.}
Recent studies have shown that scaling up computation during test-time/inference-time can effectively improve a model's reasoning performance, as exemplified by OpenAI's o1 \cite{openai2024learning,DBLP:journals/corr/abs-2407-21787,DBLP:journals/corr/abs-2408-03314,DBLP:journals/corr/abs-2408-16737}. These studies typically increase inference computation by extending the model's thinking chains or employing other techniques such as majority voting \cite{DBLP:conf/iclr/0002WSLCNCZ23}, Best-of-N with reward models \cite{DBLP:journals/corr/abs-2110-14168,DBLP:journals/corr/abs-2408-15240,DBLP:conf/naacl/YuGW24}, and tree search \cite{DBLP:conf/emnlp/HaoGMHWWH23,DBLP:journals/corr/abs-2405-03553,DBLP:journals/corr/abs-2406-03816}. Some other works train correction models to allocate more computation toward sequential corrections during test-time, thereby enhancing the model's final performance \cite{selfee2023,DBLP:conf/eacl/PaulIPBBWF24,DBLP:journals/corr/abs-2410-02884}.
In this paper, we take a different perspective by training additional critique models and delegating the responsibility for refinement to the actor itself. We investigate the significant performance gains achieved by leveraging critique models' supervision during test-time, especially for challenging problems. Motivated by these key findings during test-time, we incorporate this supervision into the exploration phase of the self-improvement process, leading to the training of stronger models.

\section{Conclusion and Future Work}

In this work, we take the preliminary step to explore how to construct high-quality and diverse training data to train critique models capable of providing step-level supervision and effective feedback without additional human annotations. By introducing critique-based supervision at test time, we demonstrate that critique models can significantly enhance the performance of reasoning models, particularly on challenging tasks. When inference computation is scaled up, critique models also yield continuous improvements and enhance the capability ceiling of reasoning models. Building on the findings from test-time, we integrate critique model supervision into the self-training process of reasoning models, proposing critique-based self-improvement and validating its effectiveness through extensive experiments. Lastly, we propose constructing step-level self-talk data based on critique data and showcasing its potential. 

Despite this, there are still limitations and future directions to work on. Specifically, (1) our method of building critique models primarily involves data construction and fine-tuning. Future work should include optimizing critique models from more perspectives, such as employing more advanced algorithms. Additionally, due to resource constraints, we are unable to conduct extensive experiments with larger-scale critique models, which may deliver superior performance and provide more reliable supervision signals. (2) We need to develop more advanced test-time scaling techniques to improve efficiency and reliability, reducing hallucinations in long thinking chains. (3) In Section \ref{sec:A Step Further: Training Step-level Self-Talk Reasoning Models via Critique Data}, we show the potential of self-talk models, and in the future, we expect to further optimize them to enhance their performance.  (4) Our work can be extended or applied to other application domains, enhancing model capabilities through reliable supervision and helpful feedback, including scientific research \citep{hendryckstest2021,DBLP:journals/corr/abs-2311-12022}, software engineering \citep{DBLP:journals/corr/abs-2107-03374,DBLP:journals/corr/abs-2108-07732}, and agentic tasks \citep{DBLP:conf/nips/Yao0YN22,DBLP:journals/corr/abs-2309-07864,DBLP:journals/corr/abs-2406-04151}. However, as we broaden the scope of applications, substantial efforts will be required to ensure safety and robustness.

\section*{Acknowledgements}
The authors would like to thank Huawei Ascend Cloud Ecological Development Project for the support of Ascend 910 processors.

\bibliography{main}

\clearpage
\appendix

\end{document}